\documentclass[pdflatex,sn-mathphys,Numbered]{sn-jnl}

\usepackage{graphicx}%
\usepackage{wrapfig}%
\usepackage{tikz}%
\usepackage{silence}
\usepackage{tabularx}
\usepackage{adjustbox}
\usepackage{support-caption}%
\usepackage{subcaption}%
\usepackage{caption}%subfigure
\usepackage{lscape}%
\usepackage{rotating}%
\usepackage{multirow}%
\usepackage{amsmath,amssymb,amsfonts}%
\usepackage{amsthm}%
\usepackage{cleveref}%
\usepackage{mathrsfs}%
\usepackage[title]{appendix}%
\usepackage{xcolor}%
\usepackage{xltabular}%
\usepackage{textcomp}%
\usepackage{manyfoot}%
\usepackage{booktabs}%
\usepackage{a4wide}
\usepackage{algorithm}%
\usepackage{algorithmicx}%
\usepackage{algpseudocode}%
\usepackage{listings}%
\usepackage{comment}%

\begin{document}
 \title[Article Title]{Exploring the Role of Convolutional Neural Networks (CNN) in Dental Radiography Segmentation: A Comprehensive Systematic Literature Review}

\author[1]{\fnm{Walid} \sur{Brahmi}}\email{bensghaierwaleed@gmail.com}
\author[1]{\fnm{Imen} \sur{Jdey}}\email{imen.jdey@fstsbz.u-kairouan.tn}
\author[1]{\fnm{Fadoua} \sur{Drira}}\email{fadoua.drira@enis.tn}

\affil*[1]
{\orgdiv{Research Groups in Intelligent Machines (REGIM Lab), University of
Sfax, National Engineering School of Sfax (ENIS), BP 1173, 3038 Sfax,},  \country{Tunisia}}

 \vspace{-1cm}\abstract{
In the field of dentistry, there is a growing demand for increased precision in diagnostic tools, with a specific focus on advanced imaging techniques such as computed tomography, cone beam computed tomography, magnetic resonance imaging, ultrasound, and traditional intra-oral periapical X-rays. Deep learning has emerged as a pivotal tool in this context, enabling the implementation of automated segmentation techniques crucial for extracting essential diagnostic data. This integration of cutting-edge technology addresses the urgent need for effective management of dental conditions, which, if left undetected, can have a significant impact on human health. The impressive track record of deep learning across various domains, including dentistry, underscores its potential to revolutionize early detection and treatment of oral health issues.\\
\textbf{Objective:}
Having demonstrated significant results in diagnosis and prediction, deep convolutional neural networks (CNNs) represent an emerging field of multidisciplinary research. The goals of this study were to provide a concise overview of the state of the art, standardize the current debate, and establish baselines for future research.\\
\textbf{Method:}
In this study, a systematic literature review is employed as a methodology to identify and select relevant studies that specifically investigate the deep learning technique for dental imaging analysis. This study elucidates the methodological approach, including the systematic collection of data, statistical analysis, and subsequent dissemination of outcomes.\\
\textbf{Results:}
In incorporating 45 studies, we identified selection criteria and research objectives, addressing significant gaps in the existing literature. These studies assist clinicians in examining dental conditions and classifying dental structures, including caries detection and the identification of various tooth types. We evaluated model performance, addressing the identified gaps, using diverse metrics that we strive to list and explain.\\ \\
\textbf{Conclusion:}
This work demonstrates how Convolutional Neural Networks (CNNs) can be employed to analyze images, serving as effective tools for detecting dental pathologies. Although this research acknowledged some limitations, CNNs utilized for segmenting and categorizing teeth exhibited their highest level of performance overall.
}




\maketitle
\section{Introduction}\label{sec1}
Medical information plays a vital role in healthcare by aiding healthcare professionals in making precise diagnoses, delivering effective treatments, and informed decisions about patient care \cite{mogli2009medical, pickering2013data}. Within the realm of medical information, medical imaging stands out as an invaluable source.

In modern healthcare, medical imaging has become an essential tool, offering crucial visual insights into the internal structures and conditions of the human body \cite{abedalla2021chest}. Within dentistry, dental imaging refers to the use of various imaging techniques to capture detailed images of the oral and maxillofacial structures. These images are invaluable for diagnosing dental conditions, planning treatments, and monitoring oral health. Dental imaging plays a vital role in enabling dental professionals to visualize and evaluate the teeth, jaws, and supporting structures.

Dental images can be acquired through different modalities, including X-rays, computed tomography (CT), magnetic resonance imaging (MRI), and intraoral cameras. Each modality offers specific advantages and is selected based on the specific diagnostic requirements and clinical scenario. 

Deep learning, in particular convolutional neural networks (CNNs) and computer vision, has become a revolutionary area of dentistry. Thanks to the use of these cutting-edge technologies, the dental industry is undergoing a period of transformation in which the analysis of dental images and the integration of artificial intelligence are revolutionizing many aspects of dental care \cite{khanna2017artificial}. By harnessing the capabilities of deep learning algorithms and computer vision techniques, dentistry has the potential to dramatically improve diagnostic performance, streamline treatment planning processes and, ultimately, improve patient outcomes.

The utilization of CNNs in dentistry encompasses the detection of structures like teeth and bone, as well as the identification of pathologies such as caries and apical lesions. Additionally, CNNs are employed to segment images by isolating the areas of interest and classify them based on specific features like enamel caries lesions or cysts. However, it is important to acknowledge the limitations within this field. One of the significant weaknesses lies in the relatively small and private nature of the available datasets, which restricts the diversity and size of the training data. Moreover, some of the AI solutions developed in this domain may lack robustness and stability, raising concerns about their reliability and performance.

To analyze dental images for tasks like lesion detection, age or sex determination, and human identification, tooth segmentation plays a vital role. In the field of oral medicine, the automatic segmentation of teeth in panoramic radiographs is a significant focus of research in image analysis. However, segmenting teeth in panoramic radiographs presents challenges due to the presence of other anatomical structures such as the chin, spine, and jaws.

Within the existing literature, a multitude of studies have employed convolutional neural networks (CNNs) for tooth segmentation and identification in dental images. These advanced deep learning techniques have exhibited promising outcomes by leveraging the capabilities of CNNs to extract meaningful features and accurately segment teeth in dental images.

This paper is organized as follows: Section \ref{sec2} provides a summary of the current state of the art in automatic tooth segmentation and references relevant surveys in the field. In Section \ref{sec3}, we delve into the principles of deep learning, exploring its various categories, common applications, and specifically zooming in on the architecture of Convolutional Neural Networks (CNNs) within the context of medical image analysis in computer vision. The subsequent sections \ref{sec4}, \ref{sec5}, and \ref{sec6} present the study methodologies, research topics, and synthesis findings, offering detailed insights into the conducted research. Finally, Sections \ref{sec7} and \ref{sec8} discuss the study's limitations and provide a conclusion.
\section{Related work}\label{sec2}
In recent years, several surveys addressing deep learning-based dental image segmentation have been conducted. This section provides a summary of related studies, highlighting the key distinctions between this study and existing surveys.

Schwendicke et al. (2019) \cite{schwendicke2019convolutional} conducted a survey on the application of Convolutional Neural Networks (CNNs) in dental and oral medicine imagery. Their review encompassed 36 relevant articles and conference proceedings spanning from 2015 to 2019, addressing clinical challenges in areas such as general odontology, cariology, endodontics, periodontics, orthodontics, dental radiology, forensic odontology, and general medicine.

Hwang et al. (2019) \cite{hwang2019overview} published a survey on deep learning in oral and maxillofacial radiology, identifying 25 pertinent papers up to December 2018 using PubMed, Scopus, and IEEE Explore databases. The study collected data on deep learning architecture, training dataset size, evaluation results, advantages and disadvantages, study objectives, and imaging modalities.

Kang et al. (2020) \cite{kang2020application} reported on a prior study focusing on the application of deep learning algorithms in dentistry and implantology. They analyzed 62 articles from MEDLINE and IEEE Xplore, categorizing them into tooth detection, numbering, segmentation, and bone segmentation. The study included articles in all languages and published before October 24, 2019, providing details on Author Year, Architecture, Input, Output, and Performance metrics.

Prados-Privado et al. (2020) \cite{prados2020dental} conducted a systematic study visualizing the state of artificial intelligence in dental applications, covering the detection of teeth, caries, filled teeth, crowns, prostheses, dental implants, and endodontic treatments. Utilizing three digital databases (PubMed, IEEE Xplore, and arXiv.org), they identified 18 relevant papers. Notably, the study did not include any segmentation methods based on deep learning.

Table \ref{tab1} summarizes the central distinguishing factors between this review and existing related work. Our contributions are outlined as follows:
\begin{itemize}
\item We delve into a comprehensive exploration of Convolutional Neural Networks (CNN), the preeminent deep learning algorithm. This involves an in-depth explanation of concepts, theory, and contemporary architectures.
\item Our study thoroughly examines crucial challenges in Deep Learning, including the shortage of training data, data imbalance, and sample data quality. We also delve into proposed solutions to address these challenges.
\item We are trying to categorize a comprehensive list of medical imaging applications using deep learning based on specific tasks.
\end{itemize}

\begin{table}[ht]
\centering
\caption{Comparison of this study with existing literature reviews on deep learning for dental image segmentation.}
\label{tab1}
\begin{tabular}{|p{1cm}|p{1cm}|p{2.3cm}|p{8.5cm}|}
\hline
{\textbf{Study}}&{\textbf{Release year}}&{\textbf{Total number of papers}}& {\textbf{Limitations}}\\ \hline
\cite{schwendicke2019convolutional}&2019& 36& Study includes studies that do not process dental images with deep learning \newline The elements impacting the model's performance were not addressed \\ \hline
\cite{hwang2019overview}& 2019 &25 &The elements impacting the model's performance were not addressed  	\\ \hline
\cite{kang2020application}& 2020 &62 & The study includes studies that do not deal with dental images.\newline The elements impacting the model's performance were not addressed\\ \hline
\cite{prados2020dental} & 2020& 18 & The article does not deal with the segmentation task.\newline The elements impacting the model's performance were not addressed\\ \hline
\end{tabular}
\end{table}
\section{Background}\label{sec3}
\subsection{Deep Learning}

\begin{figure}[ht]
\centering       
\includegraphics[width=0.43\linewidth]{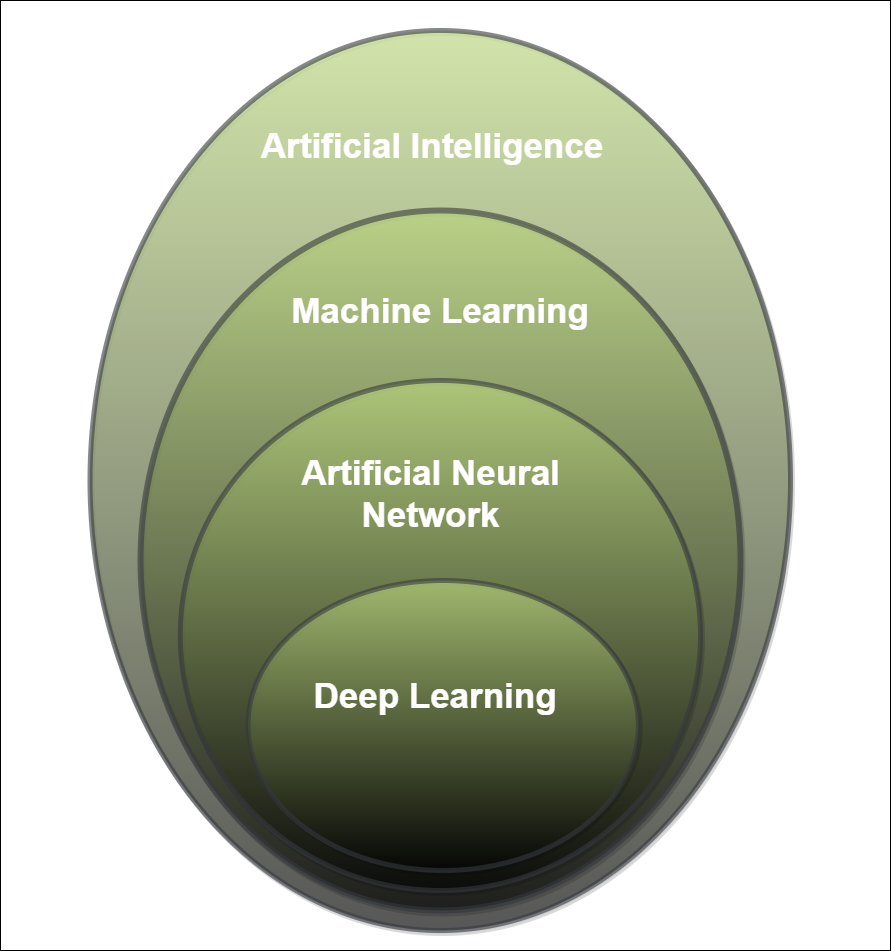}
        \caption{The Subsets of Artificial Intelligence.}
        \label{fig1}
\end{figure}

Deep learning (DL) holds significant potential to advance applications with a real impact on the field of dentistry. Positioned within Machine Learning (ML) as illustrated in Figure \ref{fig1}, DL empowers machines to emulate human intelligence in increasingly sophisticated ways \cite{bib2, bib3}. To assess and distill knowledge from extensive datasets, DL employs multiple layers of non-linear units. In a basic scenario, there are two sets of neurons: one set receives input signals, and the other set sends output signals. Upon receiving an input, the input layer modifies it before passing it on to the next layer. While the layers in a deep network may not be neurons, conceptualizing them as such proves beneficial. This approach enables the algorithm to leverage multiple processing layers comprising diverse linear and non-linear transformations \cite{bib4}.
Deep learning is part of the broader family of machine learning techniques centered around learning data representations. An observation, such as an image, can take various forms, including a set of edges, specific shape areas, or, more abstractly, a vector of intensity values per pixel.

\begin{figure}[ht]
\centering     
\fbox{
\includegraphics[width=0.7\linewidth]{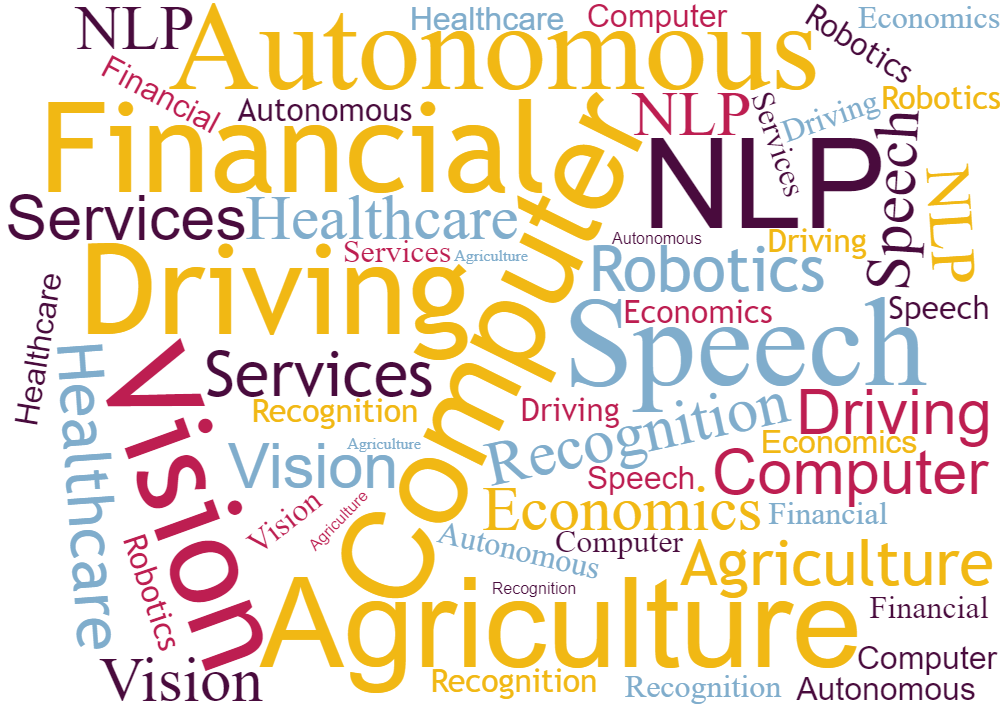}}
        \caption{Common Deep Learning Applications.}
        \label{fig2}
\end{figure}

Several deep learning architectures, such as deep neural networks (DNN), convolutional neural networks (CNN), recurrent neural networks (RNN), and others, have found application in a wide array of fields. As depicted in Figure \ref{fig2}, deep learning is frequently employed by partnerships to address diverse and challenging issues falling into several categories, including:
\begin{itemize}
    \item \textbf{Computer Vision:} Deep learning is widely used in computer vision tasks such as image classification, object detection, image segmentation, and facial recognition \cite{szeliski2022computer}.
    \item \textbf{Natural Language Processing (NLP):} Deep learning models are used to process and understand human language, including tasks such as language translation, sentiment analysis, text generation, and question answering \cite{bharadiya2023comprehensive}.
    \item \textbf{Speech Recognition:} Deep learning techniques are employed in speech recognition systems to convert spoken language into written text. This technology is used in virtual assistants, transcription services, and voice-controlled devices \cite{roger2022deep}.
    \item \textbf{Healthcare:} Deep learning is used in medical imaging analysis, disease diagnosis, drug discovery, and personalized medicine. It helps in improving accuracy, efficiency, and speed in various healthcare applications \cite{yu2023popular,egger2022medical}.
    \item \textbf{Agriculture:} Finding problematic environmental conditions \cite{ahmad2023survey}.
    \item \textbf{Financial Services:} Deep learning is employed in fraud detection, risk assessment, algorithmic trading, credit scoring, and customer behavior analysis in the financial industry \cite{huang2020deep}.
    \item \textbf{Economics:} Deep learning is employed in the field of economics to analyze and comprehend intricate economic phenomena, predict economic behavior, analyze economic data, and evaluate the effectiveness of policy interventions \cite{zheng2023deep} \cite{bouzidi4686032vision}.
    \item \textbf{Robotics:} Deep learning techniques are utilized in robotics for tasks such as object recognition, grasping and manipulation, navigation, and human-robot interaction \cite{soori2023artificial}.
    \item \textbf{Autonomous Driving :}Significant improvements in the field of autonomous driving have been made possible by the development of sensing, perception, signal processing technologies and deep learning approaches. This has reduced the effort required of human drivers and enhanced the safety of autonomous driving \cite{muhammad2020deep}.
    \item \textbf{Additional uses:}
These days, deep learning is applied in nearly every industry. There are other further deep learning uses, including automated text creation \cite{narayan2022deep}, game play \cite{hazra2022applications}, and picture captioning \cite{xiao2022new}.
\end{itemize}

\subsubsection{Convolutional Neural Networks (CNNs)}
\begin{figure}[ht]
\centering
\includegraphics[width=0.9\textwidth]{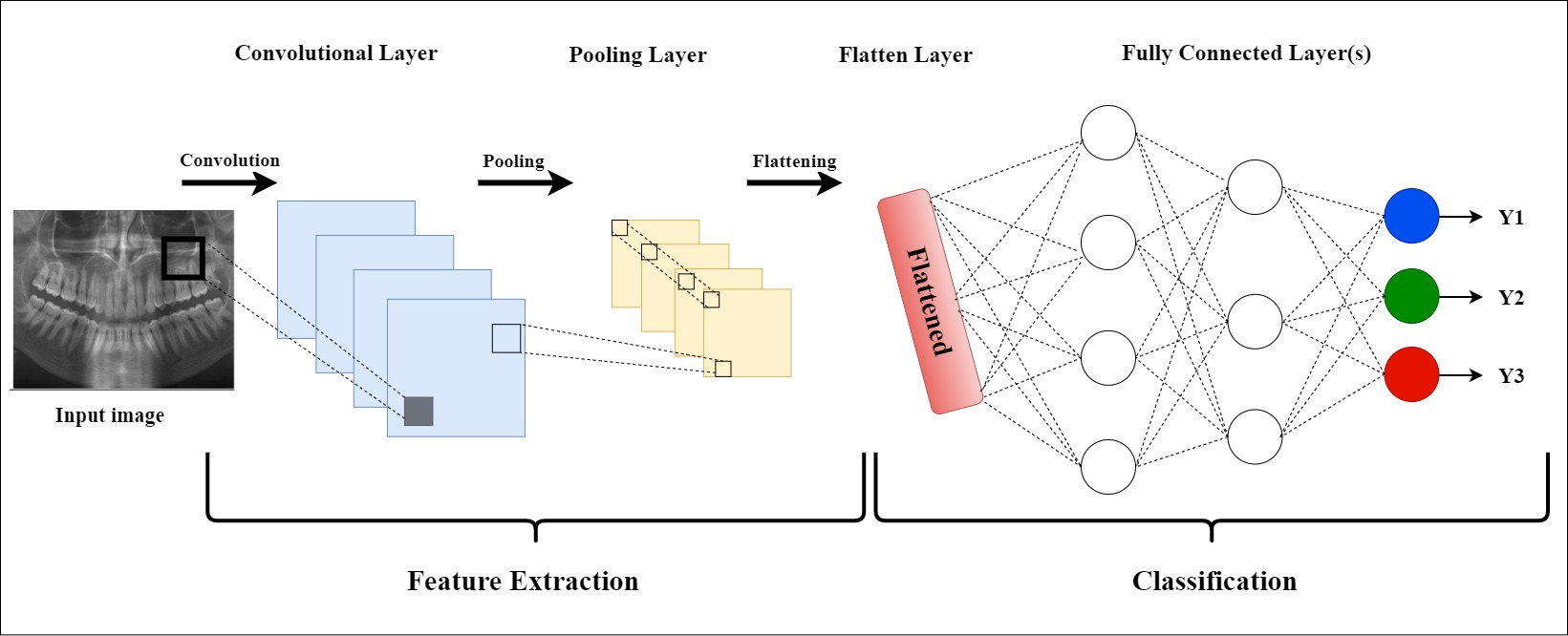}
\caption{Outline of CNN.}
\label{fig3}
\end{figure}

A Convolutional Neural Network (CNN) is a specialized deep learning model designed for processing data with a grid pattern, particularly images. Comprising multiple layers, each with a distinct purpose, the CNN architecture includes Convolutional layers, Rectified Linear Unit (ReLU) layers, pooling layers, and a fully connected layer, as depicted in Figure \ref{fig3}. CNNs automate feature extraction from images, eliminating the need for manual intervention. The CNN architecture is bifurcated into two segments: convolution and densely connected.

In convolutional layers, tensors known as feature maps are employed. A color image, represented as a 3-dimensional tensor with three channels (red, green, and blue - RGB), is denoted as (height, width, channels). To illustrate the theoretical aspects of this network type, consider the example presented in Figure \ref{fig3}, using a grayscale image of Width*Height pixels as input data. This image corresponds to a shape tensor (Width, Height, 1), indicating the number of neurons.

If Width=28 and Height=28, the tensor would consist of 784 neurons. For a color image, the tensor shape would be (28, 28, 3), resulting in 2352 neurons. It is important to note that the image, essentially a matrix where pixel colors range from 0 to 255, is normalized to a color range between 0 and 1 before entering the network.

In a typical CNN architecture \cite{alzubaidi2021review}, multiple convolution layers and a pooling layer are repeated several times, followed by one or more fully connected layers. The process of transforming input data into output data through these layers is known as forward propagation. The initial two layers, convolution, and pooling are responsible for feature extraction. The convolution layer applies filters to the input data, capturing distinct features or patterns. The pooling layer then reduces the dimensionality of the feature maps, retaining essential information while discarding some spatial details.

The final layer, the fully connected layer, maps the extracted features to the ultimate output, such as classification. It connects every neuron in the previous layer to the neurons in the current layer, allowing the network to learn intricate relationships and make predictions based on the extracted features.
\subsubsection{Categorization of Deep Learning methods}
\begin{figure}[ht]
\centering
\includegraphics[width=1\textwidth]{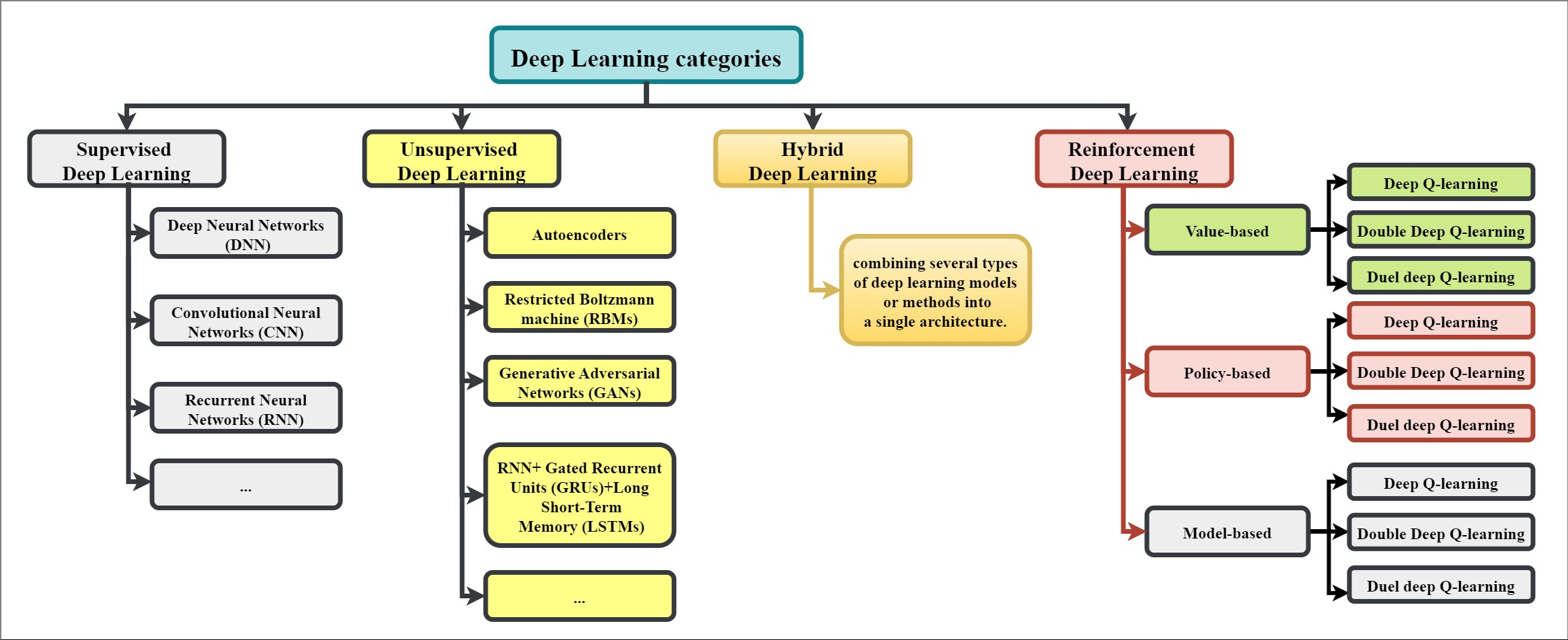}
\caption{Taxonomy of Deep Learning models.}
\label{fig4}
\end{figure}

\begin{figure}[ht]
 \centering
     \begin{subfigure}[b]{0.44\textwidth}
         \centering
    \includegraphics[height=3cm, width=\linewidth]{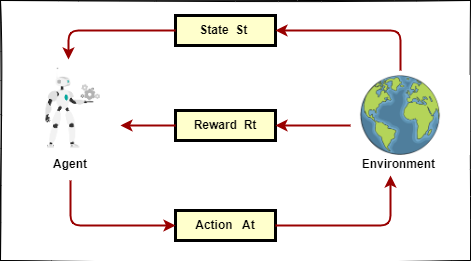}
          \caption{Reinforcement Learning Cycle }
         \label{fig5(a)}
     \end{subfigure}
\hfill
    \begin{subfigure}[b]{0.44\textwidth}
         \centering
         \includegraphics[height=3cm, width=\linewidth]{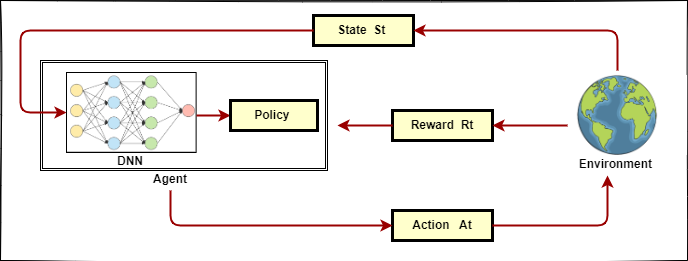}
          \caption{Reinforcement Deep Learning Cycle}
         \label{fig5(b)}
     \end{subfigure}
        \caption{Comparison of Reinforcement Deep Learning Cycle and Reinforcement Learning Cycle}
        \label{fig5}
\end{figure}

Similar to the field of machine learning, deep learning encompasses a diverse range of methodologies aimed at deriving meaningful insights from data. As figure \ref{fig4} illustrates, deep learning techniques fall into four main categories \cite{talaei2023deep, alzubaidi2021review}: supervised learning, unsupervised learning, semi-supervised learning (also known as hybrid learning), and reinforcement learning.
\paragraph{Supervised Deep Learning}
Supervised deep learning involves labeled data being used to train a model, with each data point assigned a label or target value. The model learns from labeled data by adjusting internal parameters to reduce the difference between expected and actual labels. The ultimate aim of supervised learning is for the model to generalize and predict the labels of previously unknown data points. A second test data set is used to assess the performance of the trained model. This test data set contains information that was not used during training but has known true labels. Several supervised learning approaches are available for DL, including recurrent neural networks (RNNs), convolutional neural networks (CNNs), and deep neural networks (DNNs). In addition, the RNN category covers techniques such as gated recurrent units (GRUs) and long short-term memory (LSTM).
\paragraph{Unsupervised Deep Learning}
Unsupervised learning emerges as a valuable technique when labeled data is scarce or unavailable. In such scenarios, algorithms aim to identify inherent patterns or relationships within the data without relying on predefined labels. The primary focus is on acquiring essential features or representations in the absence of explicit labels, contributing to the exploration of underlying structures or relationships within the input data.
A variety of techniques are utilized in unsupervised learning, encompassing generative networks like Generative Adversarial Networks (GANs), dimensionality reduction methods such as auto-encoders (AE) and restricted Boltzmann machines (RBM) known as Gibbs distribution,  as well as clustering algorithms \cite{talaei2023deep}. These approaches play a crucial role in diverse tasks like data clustering, dimensionality reduction, and anomaly detection.
Despite the advantages associated with unsupervised learning, particularly concerning clustering – a method grouping similar data points based on specific criteria – it has its limitations. These include the incapacity to provide precise information about data categorization and the inherent computational complexity.
\paragraph{Hybrid Deep Learning}
In this method, the learning process relies on datasets that are only partially labeled. Occasionally, both generative adversarial networks (GANs) and DRL are applied similarly to this approach. One of the benefits of this approach is its capacity to reduce the necessity for a large amount of labeled data. On the flip side, a drawback of this technique is that irrelevant input features in the training data may lead to inaccurate decisions.
\paragraph{Reinforcement Deep Learning}
Reinforcement learning (RL) is the area focused on decision-making through the acquisition of effective behavior in an environment with the goal of maximizing rewards. This learned optimal behavior is a result of interactions with the environment. Within RL, an agent possesses the ability to make decisions, observe their consequences, and adjust strategies to develop an optimal policy. In the initial stages, the agent observes the current state, takes actions, and is rewarded along with the updated state. In this process figure \ref{fig5(a)}, the immediate reward and the new state hold the potential to influence adjustments to the agent's policy. This iterative sequence continues until the agent's policy gradually approaches proximity to the optimal policy.
DRL (Deep Reinforcement Learning) addresses the primary limitations of RL, such as extended processing time required to achieve an optimal policy, thereby introducing new possibilities within the DRL framework. Broadly depicted in figure \ref{fig5(b)}, DRL leverages the characteristics of deep neural networks to enhance the learning process, resulting in improved speed and algorithm performance. In DRL, deep neural networks maintain the internal policy of the agent during interactions with the environment, determining the next action based on the current state.
DRL encompasses three main methods: value-based, where the agent learns state or state-action values and acts based on the best action in the given state; policy-based, which aims to discover an optimal policy, whether stochastic or deterministic, for better convergence in high-dimensional or continuous action spaces; and model-based, which focuses on learning the functionality and dynamics of the environment from previous observations, often using a specific model. In value-based and policy-based methods, exploration of the environment is a crucial step.
Model-based DRL involves updating the model and replanning the process. Instances of model-based DRL include imagination-augmented agents, model-based priors for model-free approaches, and model-based value expansion. While these methods are efficient when a model is available, challenges may arise with large state spaces. In such cases, the model in model-based DRL is frequently updated, and the process is replanned to achieve optimal outcomes \cite{talaei2023deep}.
\subsection{Medical Images 
 Analysis}
In the field of medical image analysis, computer vision tasks such as image classification, medical image segmentation, and object detection and recognition are prevalent. Deep learning (DL) approaches, particularly those based on convolutional neural networks (CNN), have shown superior performance in handling these challenging problems across various medical applications. In this section, we provide a comprehensive and technical evaluation of current research that leverages the latest advancements in DL and CNN-based algorithms to advance medical image analysis and understanding.

\subsubsection{Object detection}
Object detection has garnered significant interest among researchers in the past few decades. With the remarkable progress in deep learning techniques, the integration of artificial intelligence in healthcare, where object detection plays a crucial role, has become more prominent.

Object detection refers to the identification and localization of instances of specific objects within an image. This technique utilizes computer vision and image processing to recognize and locate objects in still frames and videos. Object detection systems accurately determine the number of objects present in a given area, track their positions, and classify them.

The process of object detection involves bounding box annotation, where objects are outlined with bounding boxes. The task revolves around recognizing objects captured in an image, with each object category possessing distinct characteristics for accurate categorization. Automated medical imaging can effectively identify bone fractures, abnormal cellular activity, and other medical conditions.

In the field of dentistry, several models have been developed to identify various dental features, such as tooth decay detection \cite{bib22,Lee2021DeepLF, Cantu2020DetectingCL, Lee2018DetectionAD}, periapical lesion detection \cite{Setzer2020ArtificialIF}, and dental plaque detection \cite{You2020DeepLD}. Common architectures utilized by researchers in the reviewed papers include Faster R-CNN \cite{bib29,Chen2019ADL}, SSD \cite{bib18}, and YOLO \cite{bib22,bib25,bib31}. Object detection involves two main stages: feature extraction from the target and subsequent object classification and localization.

\subsubsection{Classification}
Image classification is commonly employed to label an image or a series of images as having specific diseases or not. Traditionally, image classification involves extracting low or mid-level features to represent the image, followed by employing a trainable classifier to determine the correct label. In recent years, deep convolutional neural networks have demonstrated their superiority over manually designed low-level and mid-level features in terms of high-level feature representation. By combining feature extraction and classification networks, deep convolutional neural networks provide a unified approach that allows simultaneous training. For detailed information on DL-based medical image classification methods in clinical applications, two excellent reviews by Litjens and Ker \cite{Singha2021DeepLA, litjens2017survey} are recommended. CNNs have also found applications in dentistry for classifying medical images \cite{bib28,bib36,Chen2021HierarchicalCO}.

\subsubsection{Image segmentation}
\begin{figure}[ht]
 \centering
     \begin{subfigure}[b]{0.3\textwidth}
         \centering
    \includegraphics[width=\textwidth]{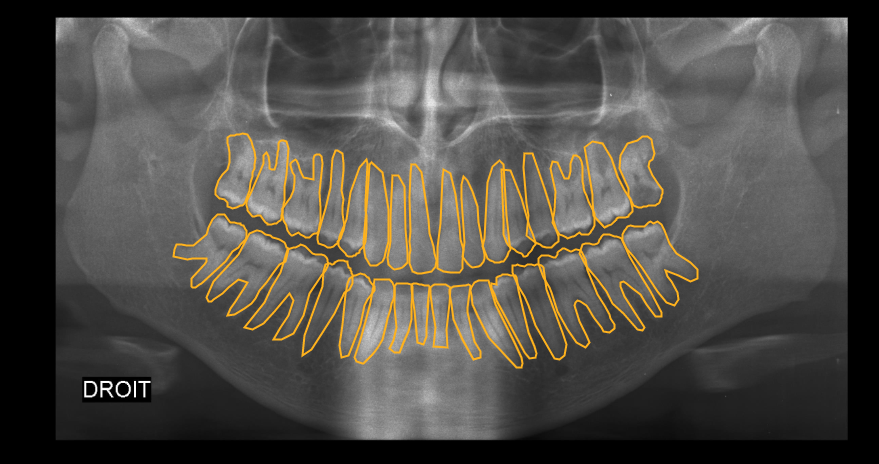}
          \caption{Semantic segmentation}
         \label{fig6a}
     \end{subfigure}
     \hfill
    \begin{subfigure}[b]{0.3\textwidth}
         \centering
         \includegraphics[height=2.35cm, width=\textwidth]{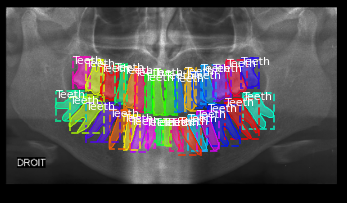}
         \caption{Instance segmentation}
         \label{fig6b}
     \end{subfigure}
     \hfill
    \begin{subfigure}[b]{0.3\textwidth}
         \centering
         \includegraphics[width=\textwidth]{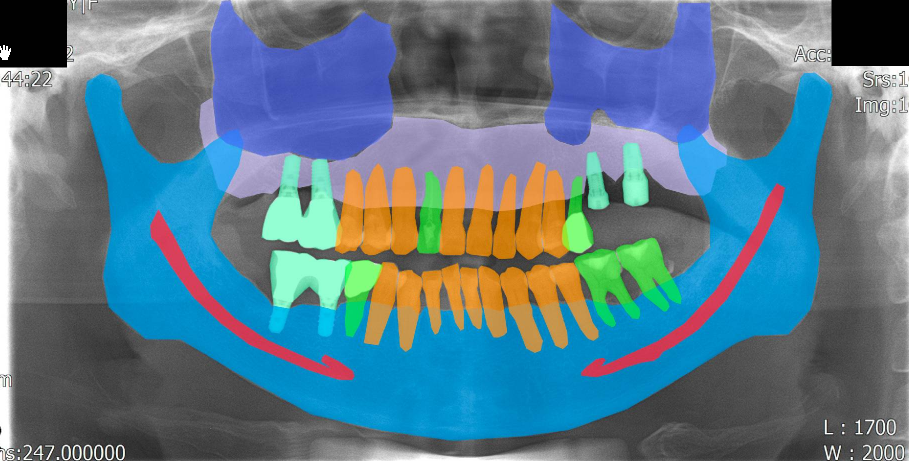}
          \caption{Panoptic segmentation}
         \label{fig6c}
     \end{subfigure}
        \caption{The three common type of image segmentation}
        \label{fig6}
\end{figure}

Image segmentation is the process of dividing an image into meaningful sub-regions or segments to identify regions of interest. It simplifies the analysis and understanding of the entire scene in further image processing stages. Segmentation involves partitioning a digital image into connected pixels or regions that share common visual characteristics, such as intensity, color, texture, histogram, or features \cite{bib10}.
As indicated in figure \ref{fig6}, there are three primary segmentation strategies commonly used in building image segmentation models: semantic segmentation \ref{fig6a}, instance segmentation \ref{fig6b}, and panoptic segmentation \ref{fig6c}. Semantic segmentation involves labeling each pixel with a class label, indicating the category it belongs to. Instance segmentation goes a step further by not only assigning class labels to pixels but also distinguishing different instances of the same class. Panoptic segmentation aims to combine the advantages of semantic and instance segmentation to provide a comprehensive understanding of the image scene.

These segmentation strategies play crucial roles in various medical image analysis tasks, enabling precise identification and delineation of regions or objects of interest, which can greatly aid in diagnosis, treatment planning, and research in dentistry and other medical fields.

\subsubsection{Others}
In the realm of medical imaging, various tasks harness deep learning techniques for enhanced outcomes. Image Reconstruction \cite{pauwels2021brief} utilizes these methods to improve the quality of medical images, enhancing resolution, reducing noise, and overall optimizing image quality. Concurrently, Image Registration \cite{kaji2020overview} involves aligning multiple medical images from different modalities or time points, aiding in image comparison for treatment planning and disease monitoring. Image De-noising \cite{kaji2020overview} employs deep learning to minimize noise and artifacts in medical images, ensuring clearer and more accurate representations. Together, these applications showcase the versatility of deep learning in advancing various aspects of medical image analysis and interpretation.
\section{Materials and methods} \label{sec4}
\subsection{Method of Study}
In this research, we employed the Systematic Literature Review (SLR) method. SLR is a rigorous and structured approach that entails assessing, interpreting, and identifying all existing research findings to address specific research questions \cite{bib13}. It adheres to a systematic process and protocols to minimize bias and ensure an objective understanding \cite{bib14}. The objective of this SLR was to identify and analyze research trends, methodologies, datasets, and frameworks pertaining to deep convolutional neural network (CNN) approaches for dental segmentation.

\subsection{Search and Selection Process}
\begin{figure}[ht]
\centering
\fbox{
\includegraphics[width=0.6\textwidth]{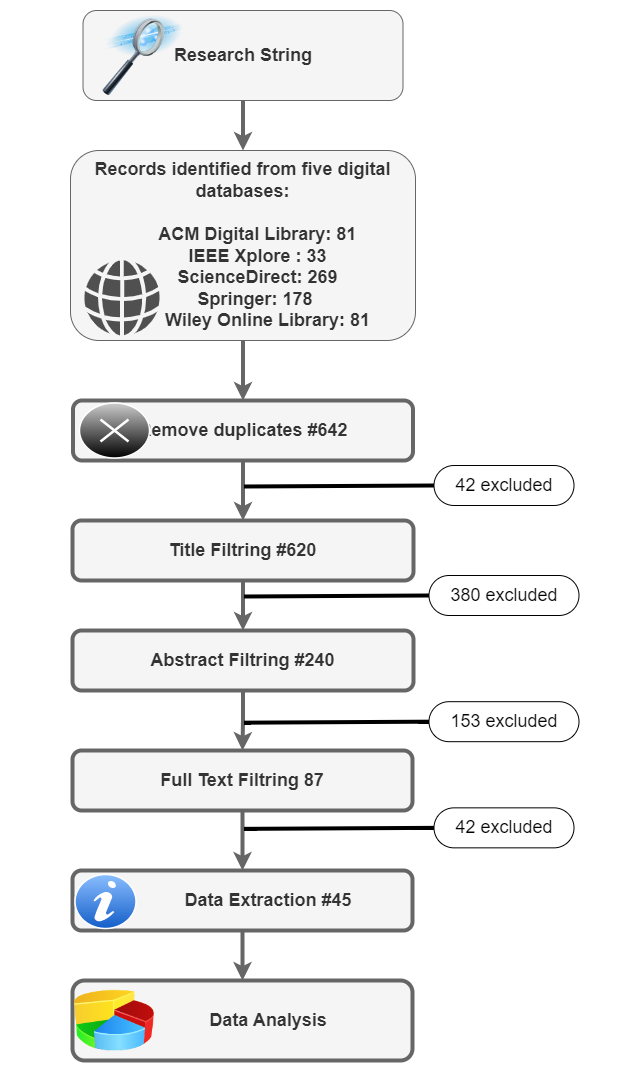}}
\caption{The systematic search and selection process.}
\label{fig7}
\end{figure}

The literature review process involved several steps, as depicted in Figure \ref{fig7}, outlining the study selection procedure for the SLR. Initially, all articles relevant to deep CNN-based approaches for dental segmentation were identified through a preliminary screening using specific search strings. Subsequently, these articles were subjected to inclusion and exclusion criteria. Following that, data extraction was conducted based on predefined criteria, and pertinent information was extracted from the selected articles. Finally, the collected data underwent analysis.

Out of a total of 642 articles initially identified, 45 articles were selected for further study based on their relevance to the research domain. The inclusion and exclusion criteria, along with the defined research protocol, were applied consistently throughout the selection process to ensure the quality and relevance of the selected articles.

By employing the SLR method, we aimed to provide a comprehensive and objective analysis of the existing research on deep CNN-based approaches for dental segmentation.
\subsubsection{Initial Search}
Before commencing the search process, we conducted an initial exploration to ensure the presence of a sufficient number of articles in the targeted field. At this stage, we identified 45 articles discussing the enhancement of dental imaging segmentation using deep Convolutional Neural Network (CNN) methods. This confirmation validated the suitability of the topic for conducting a Systematic Literature Review (SLR).

\subsubsection{Manual Search}
For the manual search, we selected appropriate search strings to capture a broad spectrum of related articles. We utilized five online databases: IEEE Xplore Digital Library \cite{ieee}, ACM Digital Library \cite{acm}, Wiley online library \cite{wiley}, ScienceDirect \cite{scd}, and SpringerLink \cite{spr}.

To mitigate false positive results, our search was confined to the titles, abstracts, and keywords of the articles. When necessary, we adjusted the search query to align with the requirements of each search engine. The search command employed was: \textbf{(("dental radiography" OR "dentistry") AND ("Deep CNN based") AND ("segmentation"))}. The manual search was conducted separately in each database, and the results were amalgamated into a comprehensive spreadsheet.

With the outlined search strategies, we identified a total of 642 relevant papers. The distribution across databases is as follows: 81 from ACM Digital Library, 33 from IEEE Xplore Digital Library, 269 from ScienceDirect Elsevier, 178 from SpringerLink, and 81 from Wiley Online Library. After eliminating duplicate articles, 45 papers proceeded to the inclusion and exclusion process (as detailed in Section 3.2).

\begin{table}[ht]
\centering
\caption{Total number of papers found in a database.}
\label{tab2}
\begin{tabular}{|p{1cm}|p{4cm}|p{6cm}|}
\toprule
\textbf{}&\textbf{Database Name}&\textbf{Total number of papers found (N)}\\
\midrule
1& ACM Digital Library& 81\\ \hline
2 & IEEE Xplore Digital Library & 33	\\ \hline
3& Sciences Direct Elsevier&269\\ \hline
4& Springer link& 178 \\ \hline
5& Willey Online Library& 81 \\ \hline
Total&\multicolumn{2}{l|}{\hspace{+4.3cm} 642} \\
\botrule 
\end{tabular}
\end{table}

\subsubsection{Selection of studies}
The study selection for our systematic literature review involved the application of predefined inclusion and exclusion criteria to the gathered data. Inclusion criteria encompassed papers published within a 10-year timeframe (2014-2023), written in English, and specifically addressing plant disease detection using deep Convolutional Neural Networks (CNN). Papers failing to meet these criteria, such as those published more than ten years ago, written in languages other than English, or not primarily focused on plant pathology detection with deep CNN, were excluded. The application of these criteria facilitated the identification of a set of relevant studies for subsequent analysis and synthesis.
\begin{table}[ht]
\centering
\caption{Inclusion and exclusion criteria. }
\label{tab3}
\begin{tabular}{|p{7cm}|p{7cm}|}
\toprule
 \textbf{Inclusion criteria}	& \textbf{Exclusion criteria}\\
\midrule
 Ten years (2014 – 2023)		& More than ten years (2013
and less)\\
\hline
English language			& Other languages\\
\hline
 Must include dental segmentation
			& Not including dental segmentation\\
\hline
 Must include Deep CNN		& Not including Deep CNN\\
\hline
Review articles	& Other types of scholarly
publication\\
\bottomrule
\end{tabular}
\end{table}

The selection process for relevant studies unfolded in three stages. Initially, screening was conducted based on the titles of the papers, with each title assessed for relevance to the study according to predefined inclusion/exclusion criteria. Subsequently, screening extended to the abstracts of the papers, where those meeting the inclusion criteria were included, and those not meeting the criteria were excluded. Finally, the same process was reiterated, this time utilizing the full text of the papers. In instances where full texts were not accessible through online databases, efforts were made to contact the authors or locate the texts from alternative sources. If the full text could not be obtained, the paper was excluded from the study.

%%%%%%%%%%%%%%%%%%%%%%%%%%%%%%%%%%%%%%%%%%
\section{Aims and Research Questions} \label{sec5}
The formulation of research questions can be organized based on four elements, known as PICOC (Population, Intervention, Comparison, Outcomes, and Context). Population refers to the targeted group of the research. Intervention is a detailed aspect of the research or issues that attract researchers. Comparison is defined as the aspect of the research where the intervention will be compared. Outcomes are the results of the intervention, and Context is the environment of the research \cite{bib15}. Before starting the search, we determined the research questions and formed the search string. According to Table \ref{tab4}, seven research questions are proposed for this study.
\begin{table}[ht]
\centering
\caption{Research Questions.}
\label{tab4}
\begin{tabular}{|p{1.7cm}|p{5.3cm}|p{8cm}|}
\toprule
\textbf{R\_Q Id}& \textbf{Research question}& \textbf{Motivations} \\
\midrule
RQ1	& What is the status of this
field of study?	& Presents a chronological summary of the studies that were selected for this SLR and shows the trend of studies in this area in recent years.\\
\hline
RQ2	& What was the key motivation for applying deep learning for dental image analysis?& Analyze the different problems addressed by deep learning algorithms in the field of dental imaging.\\
\hline
RQ3	& What deep learning implementation frameworks were used? & Understanding the technical tools and frameworks utilized in a study or project is crucial, as it provides a comprehensive insight into the realm of deep learning. \\
\hline
RQ4&What deep learning algorithms were applied?& To identify widely utilized deep learning models in dentistry, examining their applications, strengths, and potential contributions to diagnostics, treatment planning, and overall patient care. .\\
\hline
RQ5 & What evaluation metrics are most frequently used to assess a deep Convolutional Neural Network segmentation model? & Display the key ML or DL performance measures so that you may select the most appropriate ones for your use case.\\
\hline 
RQ6 &How can future studies build upon the findings of the reviewed papers to contribute novel insights or address existing limitations in the field?&  Assess the present state of the literature and recognize any deficiencies or constraints in the current research. Acknowledging these gaps creates a platform for upcoming studies to confront and surmount these challenges.\\
\hline
RQ7	& What data sources were used? & To train ML or DL models and to offer a baseline for evaluating the efficacy of the suggested architectures, numerous public datasets are presented.\\
\botrule
\end{tabular}
\end{table}
\section{Results} \label{sec6}
In this section, we expound upon the overall statistical findings derived from the encompassed primary investigations and subsequently elucidate the outcomes pertinent to the research inquiries within the subsequent sections. The roster of the 45 primary studies incorporated into this Systematic Literature Review is delineated in table \ref{tab9}.
%\subsection{What is the status of this field of study?}
\subsection{The Current Landscape of Dental Radiography Segmentation}
The 45 included studies are only in the form of journal articles. To associate the articles with their country of origin, we analyzed the affiliations of the authors of the included articles.
\begin{figure}[ht]
\centering
\fbox{
\includegraphics[width=0.7\textwidth]{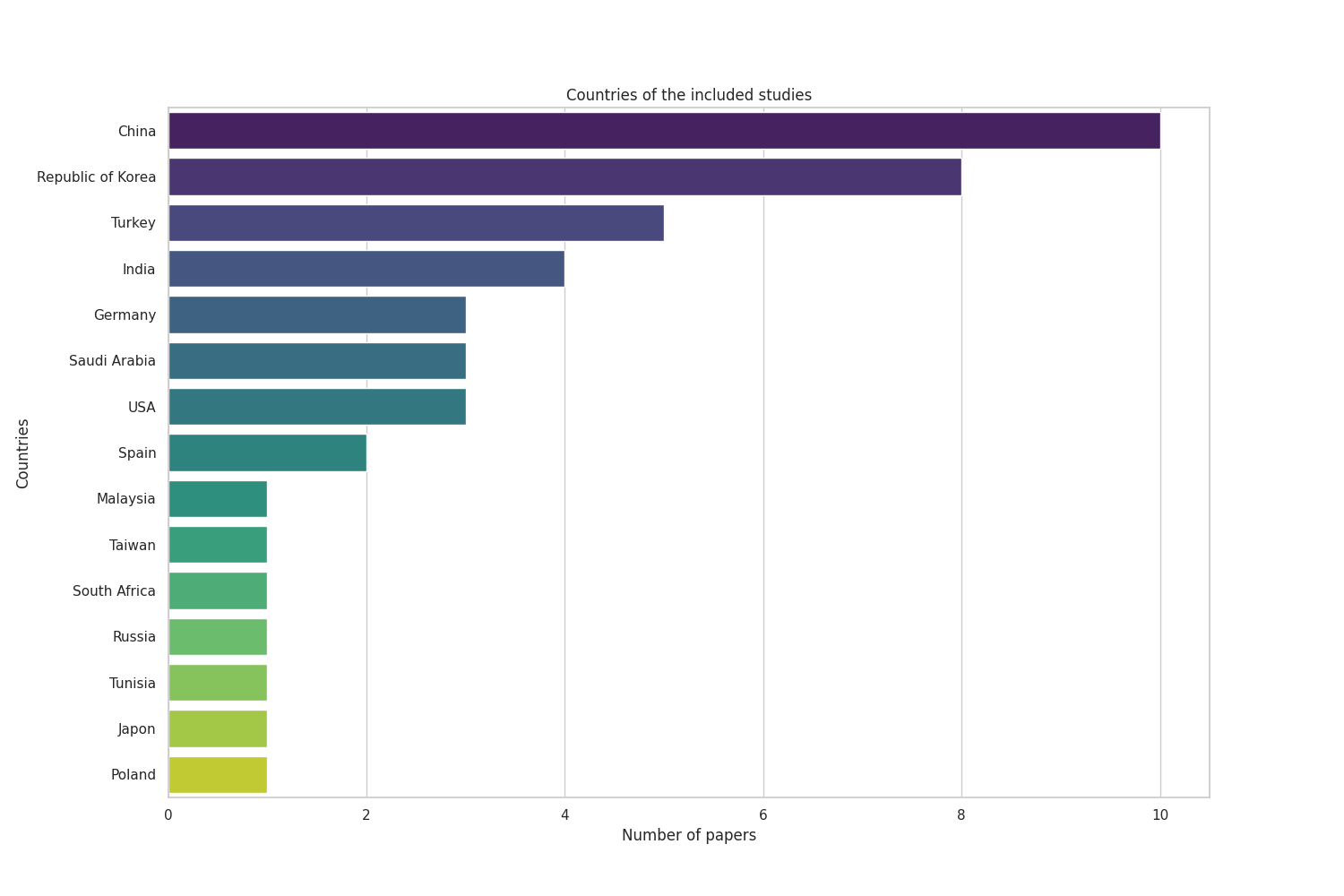}}
\caption{Countries of the included studies.}
\label{fig8}
\end{figure}  
ranking highlights the prominence of Asian countries, with China leading with 10 articles, followed by Korea with 8 articles, and India with 4 articles. Following closely are Turkey with 5 articles and the USA with 3 articles. The list also features European countries, including Germany with 3 articles and Spain with 2 articles. Additionally, other represented countries include Saudi Arabia with 3 articles, and from the Asian continent, Malaysia, South Korea, and Taiwan with 1 article each. From the African continent, South Africa and Tunisia are represented with 1 article each.

\begin{figure}[ht]
\centering
\fbox{
\includegraphics[width=0.8\textwidth]{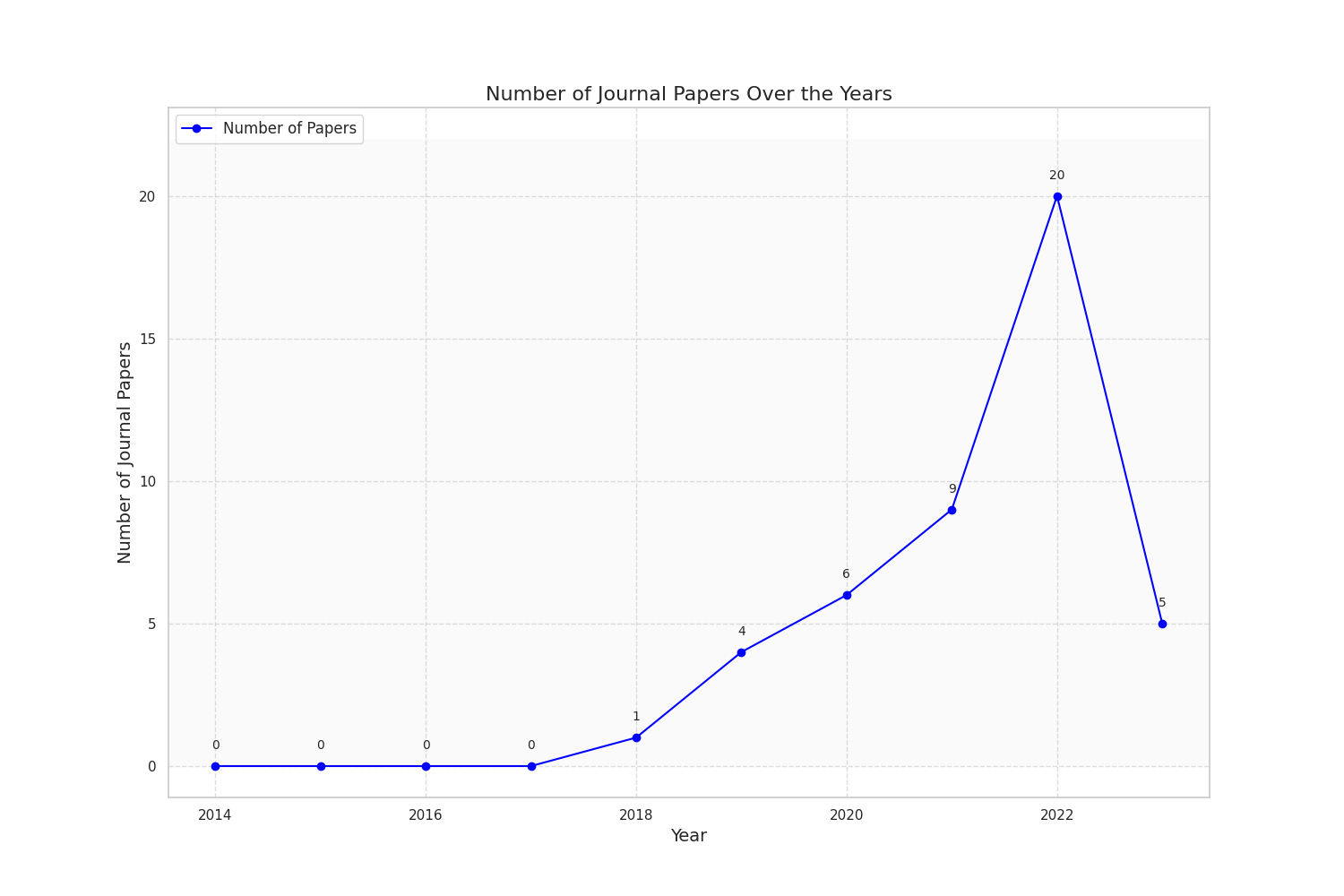}}
\caption{Yearly papers on CNNs for dental X-ray segmentation.}
\label{fig9}
\end{figure}

Figure \ref{fig9} provides an overview of the yearly growth rate of publications in this field of study. An increase in the number of publications can indicate a growing interest in the field, while a decrease could suggest that the field is a closed research area. Notably, there is a limited representation of papers from 2023, possibly due to restricted access to the most recent publications at the time of the study, indicating that the actual number of publications for that year may be higher. We observe an overall upward trend in publications, implying that although the field was a relatively unpopular research topic with no publications between 2014 and 2017, it has recently attracted considerable attention from researchers.
%\subsection{What is the purpose of applying the "deep CNN" approach to dental imaging segmentation?}
\subsection{Purposes of Deep Learning in Dental Image Segmentation}
CNNs have emerged as powerful tools in image recognition, particularly in the specialized domain of dental image segmentation. Leveraging CNN-based approaches enables rapid and accurate recognition and classification of various diagnostic processes, thereby advancing the field of dental image processing. Radiographic 2D and 3D dental images play a pivotal role in addressing a wide spectrum of dental issues.

In Table \ref{tab5}, we endeavor to outline the objectives and diseases addressed by the 45 articles in our survey.
\begin{table}[ht]
\centering
\begin{tabular}{|l|l|}
\hline
\multicolumn{2}{|c|}{\textbf{Basic Detection and Identification}} \\ \hline
Tooth detection and numbering & \cite{Chen2019ADL,  Singh2020NumberingAC, bib32, brahmi2023automatic} \\ \hline
Detection of Periapical Lesions and segmentation of cysts & \cite{bib25,Sivasundaram2021PerformanceAO,Krois2021GeneralizabilityOD,Setzer2020ArtificialIF,Ekert2019DeepLF} \\ \hline
\multicolumn{2}{|c|}{\textbf{Structural Analysis and Classification}} \\ \hline
Age and Sex determination & \cite{bib21,bib30,bib34,Rajee2021GenderCO} \\ \hline
Caries detection and segmentation & \cite{bib22,Lee2021DeepLF,Cantu2020DetectingCL,Lee2018DetectionAD,bib18,bib26,bib27,bib17} \\ \hline
Dental calculus identification & \cite{You2020DeepLD,Hsiao2021DiseaseAM} \\ \hline
Detection and classification of dental structures & \cite{bib29,bib36,Chen2021HierarchicalCO,bib23,Zhao2020TSASNetTS,Kwak2020AutomaticMC,Tian2019AutomaticCA,bib35,bib19,bib24} \\ \hline
\multicolumn{2}{|c|}{\textbf{ Advanced Condition Detection}} \\ \hline
Detecting dental conditions & \cite{bib31,bib28,Baaran2021DiagnosticCO,bib16,Chen2021DentalDD,Kheraif2019DetectionOD,bib20} \\ \hline
\end{tabular}
\caption{Automatic dental imaging segmentation purpose.}
\label{tab5}
\end{table}
The tooth numbering process assigns a unique number to each tooth using the universal tooth numbering system. In a dental panoramic image, encompassing molars, premolars, canines, and incisors in both maxilla and mandible, the maximum tooth count reaches 16 (Figure \ref{fig6a}). The distribution includes 6 molars, 4 premolars, 4 canines, and 8 incisors in each jaw.

In summary, recent advancements in deep learning, particularly convolutional neural networks (CNNs), have significantly impacted dental image processing. Researchers have successfully applied these techniques to tasks such as tooth detection and numbering, periapical lesion detection, age and sex determination, caries detection and segmentation, identification of dental structures, and diagnosis of various dental conditions. The studies highlighted demonstrate the effectiveness of CNN-based models, achieving high precision and recall rates.

Notably, these applications extend to forensic medicine, aiding in age estimation and gender identification based on dental characteristics. The growing interest in this field is evident from the increasing number of publications over the years.

Overall, the integration of deep learning approaches, as showcased in the surveyed studies, holds promise for enhancing the accuracy and efficiency of dental image analysis, contributing to improved diagnostic processes and patient care.

%\subsection{What deep learning implementation frameworks were used?}
\subsection{Popular Deep Learning Frameworks}
Continuous efforts from both the industry and the academic community have led to the development of several widely adopted deep learning frameworks. Deep learning frameworks serve as software libraries or platforms designed to streamline the development, training, and deployment of deep neural networks. A plethora of software frameworks are at our disposal for implementing Deep Learning Models (DLM), and they undergo constant updates as fresh methodologies and concepts surface. There exists a multitude of deep learning frameworks, encompassing Theano, TensorFlow, Caffe, PyTorch, MXNet, MATLAB and many others \cite{gheisari2023deep}. 

\begin{figure}[ht]
\centering
\fbox{
\includegraphics[width=0.8\textwidth]{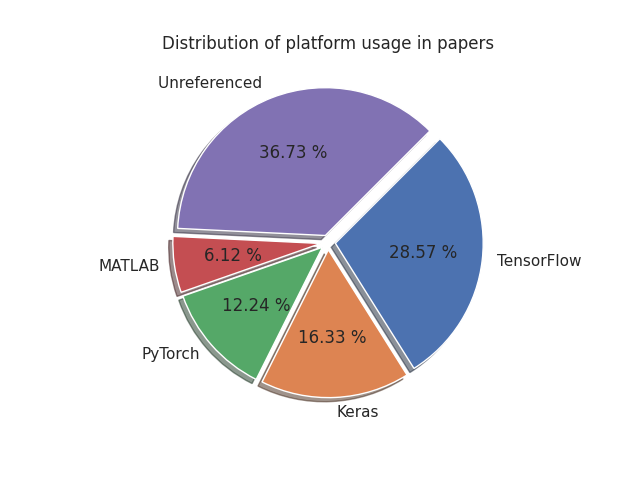}}
\caption{The distribution of papers per implementation framework.}
\label{fig12}
\end{figure}

Indeed, it is crucial to emphasize that there is no universal solution that can address every problem within the realm of machine learning. Quite often, achieving success necessitates the utilization of a combination of diverse tools and frameworks.  Drawing from the results of our comprehensive literature review, illustrated in Figure \ref{fig12}, it becomes evident that Keras, PyTorch, and TensorFlow have surfaced as the dominant libraries in the realm of research. This prominence can be attributed to the fact that both Keras and TensorFlow platforms are open-source and readily accessible tools. Researchers tend to favor these alternatives due to their lack of subscription fees, a significant departure from MATLAB, where users often encounter expenses for access.
In this section, we will examine various popular machine learning and deep learning frameworks and libraries.
\subsubsection{TensorFlow}
TensorFlow is an open-source software library designed for numerical computations using data flow graphs. It was created and is maintained by Google's Brain team in their Machine Intelligence research organization for machine learning and deep learning applications.
This library is tailored for large-scale distributed training and inference tasks. In TensorFlow, nodes in the computational graph represent mathematical operations, while the edges represent the multidimensional data arrays known as tensors that flow between these operations.  The architecture of distributed TensorFlow includes master and worker services with kernel implementations.

TensorFlow, as the most widely adopted deep learning framework \cite{wang2019various}, boasts numerous unparalleled features. However, its development remains ongoing, addressing various shortcomings. Here are some of TensorFlow's notable advantages:

\begin{itemize}
    \item Multi-language Support: TensorFlow offers support for multiple programming languages, with the potential for future expansion to include even more languages.
    \item High Performance: TensorFlow delivers excellent performance and offers robust support for various hardware configurations, including multi-CPU, GPU, or hybrid platforms.
    \item Real Portability: TensorFlow exhibits a high level of portability, surpassing many other frameworks. Its close integration of research and product development, coupled with tools like TensorFlow Serving, enables seamless translation of innovative ideas into practical products.
\end{itemize}
\subsubsection{Keras}
Keras is an open-source neural network library renowned for its simplicity and modularity. It is predominantly implemented in pure Python, making it accessible to a wide range of users. Keras was specifically designed to facilitate the process of experimenting with deep neural networks, and it offers a comprehensive toolkit for building and training these networks. Some key aspects of Keras include \cite{murad2023medical}:
\begin{itemize}
    \item Minimalist Framework: Keras is celebrated for its minimalist framework, which reduces complexity and enables users to focus on the core aspects of deep learning model development.
    \item Modularity: Its modular design allows users to build neural networks by assembling standalone, fully-configurable modules. This modularity extends to layers, loss functions, activation functions, and more.
    \item Ease of Experimentation: Keras simplifies the experimentation process, making it easier to explore various deep learning architectures and ideas.
    \item Compatibility: Keras exhibits compatibility with multiple backends or deep learning frameworks, providing users with the flexibility to select the backend that aligns best with their specific requirements and preferences. This adaptability allows users to seamlessly integrate Keras into their existing infrastructure and leverage the strengths of different frameworks for their deep learning projects.
    \item Versatility: Keras supports Convolutional Neural Networks (CNNs), Recurrent Neural Networks (RNNs), and other deep learning models. It accommodates multiple input and output structures, making it versatile for various tasks.
\end{itemize}
\subsubsection{PyTorch}
Facebook, Twitter, and Yandex have employed the Torch framework, available in two iterations: the traditional version based on the LuaJIT language, and the Python-based variant known as PyTorch, launched by Facebook in January 2017. PyTorch \cite{paskze2017tensors}, functioning as a Python package, offers two prominent features: GPU-accelerated tensor calculations and the ability to construct dynamic neural networks. These features distinguish PyTorch as a versatile and powerful tool in the realm of deep learning.\\
In contrast to many other frameworks such as TensorFlow, which exhibit a primarily static nature necessitating users to build and reuse fixed network structures \cite{wang2019various}, PyTorch stands out with its dynamic approach. It leverages Reverse-mode auto-differentiation, allowing users to modify network performance with minimal delay or computational overhead. While this technology is not exclusive to PyTorch, it boasts one of the most rapid and efficient implementations, establishing PyTorch as a prominent player among deep learning frameworks
\subsubsection{Caffe}
Caffe, developed by the University of California, Berkeley Vision and Learning Center (BVLC), is a convolutional neural network framework designed around a C++/CUDA/Python architecture. Jia Yangqing is credited as its creator \cite{jia2014caffe}. Built upon feed-forward convolutional neural network architectures, such as CNNs, Caffe offers multiple interfaces for command line, MATLAB, and Python. Notably, it excels in feature extraction and seamlessly switches between CPU and GPU, harnessing GPU acceleration for accelerated CNN learning. On April 18, 2017, Facebook introduced an enhanced iteration of Caffe known as Caffe2, operating under the BSD license agreement. With its expanded feature set, Caffe2 has the potential to surpass Caffe in popularity, emerging as a more prominent deep learning framework in the future.
\subsubsection{MXNet}
In November 2016, Amazon introduced MXNet \cite{mxnet2018flexible}, an open source, lightweight, portable, and flexible deep learning library developed by DMLC (Distributed Machine Learning Community), which allows users to mix symbolic programming mode and command programming 
mode to maximize efficiency and flexibility It quickly gained prominence as one of the most performant frameworks. Following its endorsement, Amazon expanded its services to include text-to-speech recognition and an image analysis service called Recognition.
MXNet is officially recommended by Amazon Web Services (AWS) and is notably the first framework to offer support for multi-GPU and distributed training, distinguishing it from its peers. What sets MXNet apart is its wide range of language bindings, making it accessible to developers using C++, Python, R, Julia, MATLAB, and JavaScript. MXNet represents a powerful tool in the realm of deep learning, offering efficiency, flexibility, and versatility to its users.

It is also noteworthy that many research articles did not mention the specific implementation platform used. It is essential to provide comprehensive details regarding the implementation or to include a link to the code. Unfortunately, some articles overlooked this aspect. We recommend that researchers not only specify the deep learning implementation platform but also delve into other technical aspects employed in their model development when writing their papers for academic journals.

%\subsection{What deep learning algorithms were applied?}
\subsection{Applied Deep Learning Algorithms}
In the background section, we have described Convolutional Neural Networks (CNNs), which are highly relevant for deep learning. Below, we will discuss the most commonly used types of deep learning networks that have been employed in the reviewed studies, based on the publications we have selected.
%\begin{landscape}
\begin{longtable}{|p{1.5cm}|p{0.7cm}|p{6cm}|p{2cm}|p{3.2cm}|}
\caption{Summary of Pre-Trained Models}
\label{tab}
\\
\hline
\textbf{Model}&\textbf{Year}& \textbf{Architecture}& \textbf{Main contribution}& \textbf{Limitations}
\\
\hline
LeNet \cite{lenet}&1998&Two convolutional (conv) layers for feature extraction. Two sub-sampling layers for spatial dimension reduction. Two fully connected layers for a global understanding of the input. An output layer featuring a Gaussian connection, suggesting the use of either a Gaussian activation function or Gaussian-weighted connections.&The recognition of handwritten digits.&• Inadequate adaptation to a variety of image classes.
\newline • Utilization of oversized filters
\newline • Limited capability in low-level feature extraction\\
\hline 
VGG (Visual geometry group) \cite{vgg}&2014&Multiple layers of small 3x3 convolutional filters, followed by max Pool\_ Layers. The fully connected layers use ReLU activation and dropout regularization to prevent overfitting.& The increase in the depth of a network (number of layers) is a critical factor in improving performance.& • Incorporation of computationally demanding fully connected layers.\\
\hline 
GoogLeNet (Inception-V1) \cite{googlenet} &2015&introduced the inception block in CNN. This block uses filters of different sizes to capture spatial information at various scales, aiming for high accuracy with reduced computational cost. It replaces traditional convolutional layers with these blocks, incorporating a 1x1 convolutional filter bottleneck for computation regulation. Sparse connections and global average pooling reduce redundancy, cutting parameters from 138 million to 4 million.&The development of an inception module, which significantly reduces the network's parameter count, leading to improved computational efficiency.& • Cumbersome parameter customization owing to a heterogeneous topology
\newline • Potential loss of valuable information due to a representational bottleneck
\\
\hline 
Inception V3 \cite{inceptionv3}&2015&The architecture incorporates factorized 7x7 convolutions, spatial factorization using asymmetric convolutions, and an auxiliary classifier for label information propagation. Additionally, batch normalization and ReLU activation functions are applied. With around 23 million parameters&Addresses the representational bottleneck by replacing large size filters with smaller ones. &• Elaborate architecture design
\newline • Absence of homogeneity
\\
\hline 
Faster R-CNN \cite{faster}&2015&A single-stage model with end-to-end training, utilizing a novel Region Proposal Network (RPN) for efficient region proposal generation. It features key elements such as the fully convolutional RPN, which generates proposals at low computational cost, and the sharing of convolutional features between the RPN and the detection network. Training involves stochastic gradient descent (SGD) optimization for convolution layers, RPN weights, and the last fully connected layer weights.&Region proposal network (RPN)& • Complexity
\newline • Training Time
\newline • Detection Time
\newline • Resource Intensiveness
\newline • Model Complexity
\\
\hline 
U-Net \cite{unet}&2015& A U-shaped encoder-decoder network architecture, consisting of four encoder blocks and four decoder blocks connected via a bridge. Each encoder block involves two 3x3 convolutions, followed by batch normalization and a ReLU activation function. The decoder network employs transposed convolutional layers for upsampling, aiming to restore the original image size. The inclusion of skip connections between encoder and decoder blocks facilitates the recovery of spatial information lost during downsampling. &Skip connections.&•Sensitivity to Input Size
\newline • Hardware Memory Requirements
\\
\hline 
ResNet (Residual Network) \cite{resnet}&2015&
ResNet utilizes residual blocks, which are stacked together to form the overall network. The original ResNet had 34 layers with 2-layer blocks, while advanced variants, like ResNet50, used 3-layer bottleneck blocks for improved accuracy and reduced training time. Skip connections in ResNet add outputs from previous layers to stacked layers, enabling the training of much deeper networks than was previously possible.&Residual layer.&• Complexity in architecture
\newline • Degradation of feature-map information during feedforward
\newline • Over-adaptation of hyperparameters for specific tasks due to the repeated stacking of identical modules
\newline • Numerous layers may contribute minimal or no information.
\newline • Relearning of redundant feature maps may occur.
\\

\hline
MobileNet \cite{mobilenets}&2017&The architecture comprises convolution layers, depthwise convolution layers with BN and ReLU, pointwise convolution layers with BN and ReLU, Global Average Pooling, Reshape, Dropout, additional Convolutional layers, Softmax, and Reshape. With approximately four million parameters, MobileNet is designed to be lightweight. It features two basic units: 3x3 Convolution and 3x3 Depthwise Convolution followed by 1x1 Convolution, using separable filters to reduce input channel numbers.&MobileNet's architecture is optimized for mobile and embedded devices, utilizing depthwise separable convolutions. Sequential layers include depthwise and pointwise convolutions with Batch Normalization and ReLU.&• Challenges arise on resource-constrained micro-controllers.
\newline • Sacrifices in accuracy compared to larger architectures, especially in detailed tasks.
\newline • Inherent tradeoff in accuracy due to its small size, though often minimal.
\\
\hline  
DenseNet \cite{densely}&2017&DenseNet's architecture utilizes Dense Blocks with densely connected layers to maximize feature reuse. Each block includes an input, output, and a growth rate (k) that regulates information flow to the next layer. Transition Layers connect dense blocks using 1x1 convolutions, 1x1 pooling, and 3x3 convolutions, striking a balance between model complexity and information flow. The Growth Rate (k) governs the number of additional features in each dense block, impacting information flow and network generalization. Finally, the Output Layer is typically a fully connected layer mapping input features to the desired class labels.&Densely connected architecture&• Memory Usage
\\
\hline 
YOLO (You Only Look Once) \cite{yolo}&2016&YOLO uses a fully convolutional neural network that passes the image through a series of convolutional layers to extract features. It predicts object boundaries and class labels simultaneously, making it a powerful and efficient deep learning model for object detection tasks.&Real-time object detection& • Detecting small or complex-shaped objects and accurately handling very large objects\\
 \hline 
 Mask RCNN \cite{mask}&2017&Incorporates a backbone network for feature extraction, a Region Proposal Network (RPN) for candidate region proposals, and a parallel branch for predicting segmentation masks alongside bounding box coordinates and class probabilities. This comprehensive framework excels in instance segmentation tasks, providing pixel-level masks for accurate object delineation.& RoIAlign layer.&• False alerts
\newline • Missing labels\\
\hline
\end{longtable}
%\end{landscape}

%\subsection{What evaluation metrics are most frequently used to assess a deep Convolutional Neural Network segmentation model?}
\subsection{Evaluation Metrics for Deep CNN Segmentation in Dental Imaging}
The tools or mechanisms for measuring model quality are known as evaluation metrics, a crucial component in developing effective ML or DL models. These metrics vary based on tasks, applications, and models. In the context of dental image segmentation, we outline the evaluation metrics employed in the selected papers. \\
Table \ref{tab6} presents various performance measurement formulas.
For a comprehensive overview of commonly used evaluation measures, along with their interpretation and implementation, refer to Table \ref{cm}. These metrics are derived from attributes in the Confusion Matrix (Figure \ref{tab6}), a two-dimensional matrix providing information about Actual and Predicted classes. Key attributes include:
\begin{table}[h]
  \centering
  \renewcommand{\arraystretch}{4}
  \begin{tabular}{c|c|c|}
    \multicolumn{1}{c}{} & \multicolumn{1}{c}{\textbf{Predicted Class 0}} & \multicolumn{1}{c}{\textbf{Predicted Class 1}} \\
    
    \textbf{Actual Class 0} & True Negative (TN) & False Positive (FP) \\
    
    \textbf{Actual Class 1} & False Negative (FN) & True Positive (TP) \\  
  \end{tabular}
  \caption{Confusion Matrix}
  \label{confusion_matrix}
  \label{cm}
\end{table}

\begin{itemize}
    \item \textbf{TP}  True-Positive is the number of Positive cases classified correctly.
\item \textbf{TN} True-Negative is the number of Negative cases classified correctly.
\item \textbf{FP} False-Positive is the number of cases where the model predicted Positive
class, but the actual class was Negative. Also known as Type 1 Error.
\item \textbf{FN} False-Negative is the number of cases where the model predicted Negative class, but the actual class was Positive. Also known as Type 2 Error.
\end{itemize}
\begin{landscape}
\begin{longtable}{|p{3cm}|p{8cm}|p{8cm}|p{3cm}|}
\caption{Performance Evaluation Metrics.}
\label{tab6}
\\
\hline
\textbf{Metric} & \textbf{Formula} & \textbf{Description}& \textbf{Study}
\\
\hline
\newline 
 Accuracy (Pixel Accuracy (PA))&  \newline $\frac{Number of Correct Predictions (TP\,+\,TN)}{Total Number of Predictions (TP\,+\,TN\,+\,FP\,+\,FN)}$\newline \newline $ = \frac{(Number \,of \,correctly\, classified\, pixels)}{(Total\, number\, of \,pixels)} * 100 \%$  &Is one way to determine how frequently the algorithm correctly classifies a data point. This is the number of items correctly identified as true positives or true negatives out of the total number of items. &\cite{bib22, Cantu2020DetectingCL, Lee2018DetectionAD, bib25, bib31, bib28, bib36, Chen2021HierarchicalCO, Singh2020NumberingAC, bib32, Sivasundaram2021PerformanceAO, bib21, bib30, bib34, Rajee2021GenderCO, bib26, bib27, bib17, Hsiao2021DiseaseAM, bib23, Zhao2020TSASNetTS, Tian2019AutomaticCA, bib19, bib24, bib16, Kheraif2019DetectionOD,choi2023,Chun2023,amasya2023}
\\
\hline 
\newline   Precision &  \newline  $ \frac{TP}{TP\,+\,FP}$ & Agreement between true class labels and machine predictions. It is calculated by adding all true positives and false positives in the system across all classes.&\cite{Lee2021DeepLF, bib29, Chen2019ADL, bib31, bib28, bib36, Chen2021HierarchicalCO, bib21, Rajee2021GenderCO, bib26, bib27, bib17, Zhao2020TSASNetTS, bib24, Baaran2021DiagnosticCO, Chen2021DentalDD, Kheraif2019DetectionOD,brahmi2023automatic,
choi2023,Chun2023,ali2023} \\
\hline
\newline Recall [(Sensitivity) (True Positive Rate  TPR)]&  \newline  $\frac{TP}{TP\,+\,FN}$ & The ability of a classifier to identify class labels. It is calculated by adding all true positives and false negatives in the system across all classes. & \cite{bib22, Lee2021DeepLF, Cantu2020DetectingCL, Lee2018DetectionAD, Chen2019ADL, bib25, bib31, bib28, bib36, Chen2021HierarchicalCO, Singh2020NumberingAC, bib32, Sivasundaram2021PerformanceAO, Krois2021GeneralizabilityOD, Ekert2019DeepLF, bib21, Rajee2021GenderCO, bib26, bib27, bib17, bib23, Zhao2020TSASNetTS, bib35, bib24, Baaran2021DiagnosticCO, Chen2021DentalDD, Kheraif2019DetectionOD,choi2023,Chun2023}
 \\
\hline
\newline Specifcity &  \newline  $\frac{TN}{FP\,+\,TN}$ & This is referred to as the true negative rate. This function computes the proportion of actual negative cases predicted as negative by our model. & \cite{bib22,Cantu2020DetectingCL, Lee2018DetectionAD, bib29, bib25, bib28, bib36, Singh2020NumberingAC, bib32, Sivasundaram2021PerformanceAO, Krois2021GeneralizabilityOD, Ekert2019DeepLF, bib21, Rajee2021GenderCO, bib17, bib23, Zhao2020TSASNetTS, bib35, Kheraif2019DetectionOD,Chun2023}
\\
\hline
\newline   F1-score &  \newline 
$2\,*\, \frac{\,Precision\,*\,Recall}{Precision\,+\,Recall} $  & Measures the effectiveness of identification when recall and precision are of equal importance. &\cite{Lee2021DeepLF, Cantu2020DetectingCL, bib29, bib25, bib31, bib28, bib36, Chen2021HierarchicalCO, Sivasundaram2021PerformanceAO, Krois2021GeneralizabilityOD, bib26, bib24, Baaran2021DiagnosticCO, Kheraif2019DetectionOD,brahmi2023automatic, choi2023, amasya2023} 
\\
\hline
\newline Cohen’s Kappa &  \newline $\frac{P_0\,-\,P_e}{1\,-\,P_e}$ \newline where $P_0$ is the observed relative agreement between raters and $P_e$ is the hypothetical probability of chance agreement.  & Represents the level of precision and reliability in a statistical classification and assesses the degree of agreement between two raters (judges) who each assign items to mutually exclusive categories. &\cite{bib28, bib30,choi2023,amasya2023}\\
\hline
 Dice similarity coefficient (DSC)& $\frac{(2 \,* \,area \,of\, overlap)}{(2 \,* \,area \,of \,overlap)
(Total \,number\, of\, pixels \,in \,both images),}$  & Used to assess how well a predicted segmentation matches the corresponding ground truth in terms of pixel-level agreement.&\cite{Setzer2020ArtificialIF, bib25, bib32, Sivasundaram2021PerformanceAO, Rajee2021GenderCO, bib26, bib23, Zhao2020TSASNetTS, bib16} \\
\hline
 Receiver operating characteristic (ROC)&  Doesn't have a specific formula &  ROC curve is graphical display of sensitivity (TPR)
on y-axis and (1 – specificity) (FPR) on x-axis for
varying cut-off points of test values. &\cite{Lee2018DetectionAD, bib18, bib35,Chun2023} \\
\hline
 Area under the curve (AUC) Average precision (AP))& AUC = $ 1-\frac{1}{2}(\frac{FP
}{FP + TN}+\frac{FN
}{FN + TP})$ \newline \newline 
AP =$\int_{r=0}^{1} p(r) \, \mathrm{d}r$  \newline\newline mean Average precision (mAP)
\newline mAP =$\frac{1}{N}\sum_{k=1}^{K=n}AP_k$ & As a single scalar value, AUC measures the performance of a binary classifier. It lies between [0.5–1.0], where the minimum value represents random classification and the maximum value represents perfect classification. &\cite{bib22, Lee2018DetectionAD, Chen2019ADL, Chen2021HierarchicalCO, Ekert2019DeepLF, bib34, bib27, Hsiao2021DiseaseAM, bib35, Chen2021DentalDD, brahmi2023automatic,Chun2023,ali2023} 
\\
\hline
Intersection over Union (IoU) [Jaccard index (JI)]&$\frac{Area_{pred}\, \bigcap \, Area_{gt}}{Area_{pred}\, \bigcup\,Area_{gt}}$ \newline $=\frac{TP}{TP + FP + FN}
$& Used to evaluate the performance of object detection by comparing the ground truth bounding box to the preddicted bounding box. &\cite{You2020DeepLD, bib29, Chen2019ADL, bib25, bib32, Sivasundaram2021PerformanceAO, bib19, bib16, Chen2021DentalDD, bib20}
\\
\hline
\newline Matthews correlation coefficient (MCC)& \newline $\frac{\text{TP}\cdot\text{TN}-\text{FP}\cdot\text{FN}}{\sqrt{ (\text{TP}+\text{FP})\cdot(\text{TP}+\text{FN})\cdot(\text{TN}+\text{FP})\cdot(\text{TN}+\text{FN}) }} $ & The MCC is used for binary classification and is considered particularly useful in unbalanced class situations. takes values between 1 (perfect correlation between ground truth and predicted outcome) and -1 (inverse or negative correlation) - a value of 0 denotes a random prediction. &\cite{Cantu2020DetectingCL} \cite{bib25} \cite{bib28}  \\\hline 

%\newline\newline Volumetric overlap error (VOE)& \newline \newline  $1-\frac{V_{gt}\, \bigcap \, V_{gt}}{V{gt}\, \bigcup\,V{pred}}$ &
%Is the rate between intersection and union of two sets of segmentation. Employed to assess the concordance between two volumetric regions or structures. It quantifies the dissimilarity by measuring the extent of non-overlapping volume between a segmented or predicted volume and a reference or ground truth volume. The lower the VOE, the better the agreement and overlap between the two volumes, making it a valuable tool for evaluating the accuracy of volumetric segmentation.&  \\\hline
%\newline\newline Relative volume difference (RVD)& \newline\newline\newline\newline \newline &
%Used to quantify the percentage difference between the measured or estimated volume of an object and its true or reference volume.&  \\\hline
\end{longtable}
\end{landscape}

In the field of evaluating dental image segmentation models, it is important to note that, in addition to the mentioned metrics, some papers also utilize specific measures such as Volumetric Overlap Error (VOE) and Relative Volume Difference (RVD) \cite{Chun2023}. These measures provide a thorough assessment of model performance, especially when applied to volumetric tomographic images like Cone Beam Computed Tomography (CBCT).  Volumetric Overlap Error (VOE) measures the degree of mismatch between segmented regions and actual regions, offering insights into the accuracy of segmentation in terms of spatial localization. Relative Volume Difference (RVD) quantifies volume differences between segmented regions and true anatomical volumes, providing an assessment of accuracy in terms of quantity. 
Thus, by considering these additional metrics, researchers can obtain a more comprehensive evaluation of the performance of dental image segmentation models, particularly when applied to volumetric tomographic images such as CBCT scans.

\begin{figure}[h]
\centering
\fbox{
\includegraphics[width=0.8\textwidth]{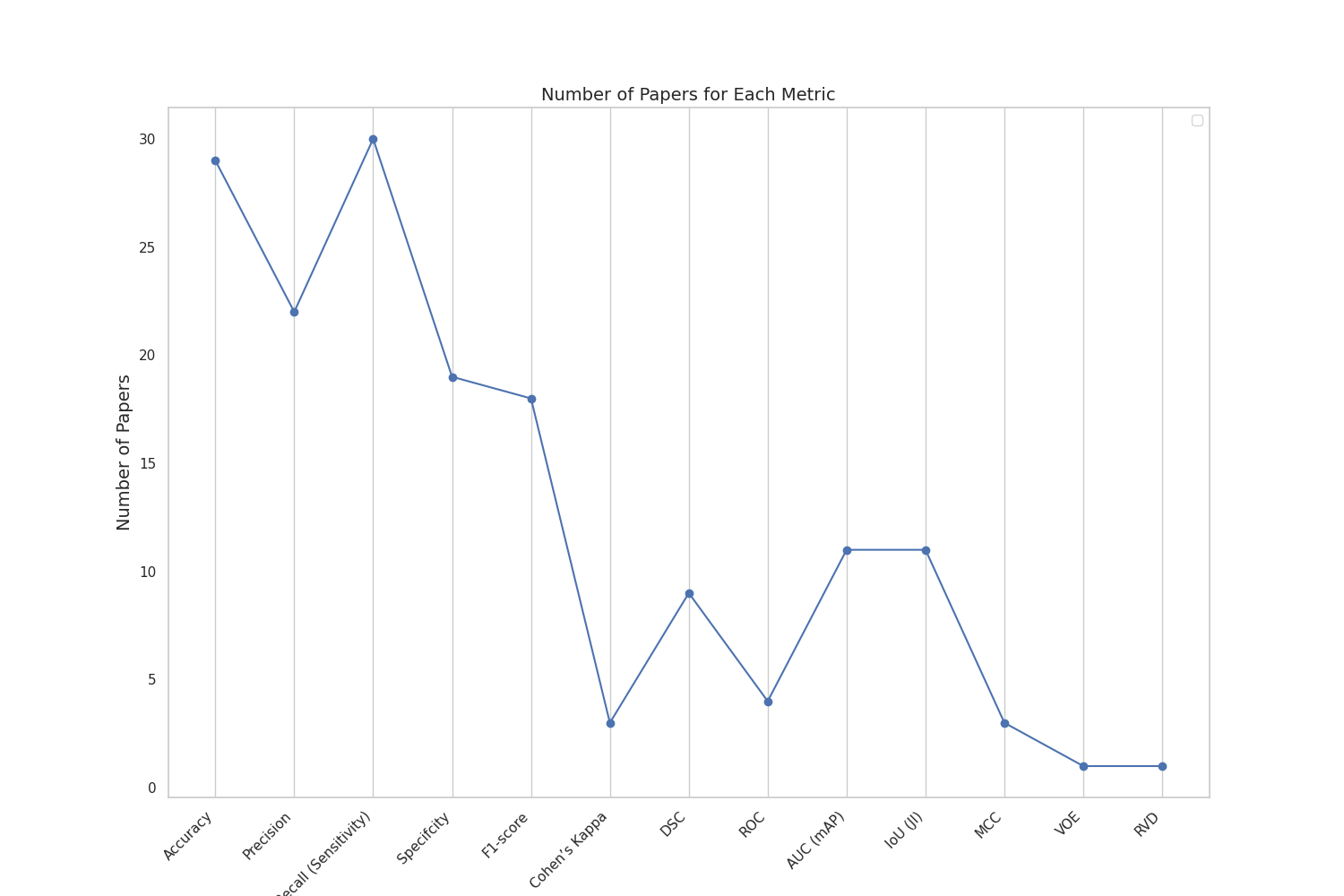}}
\caption{The summary of Performance evaluation measures reviewed in this SLR.}
\label{fig11}
\end{figure}

As shown in Figure \ref{fig11}, and according to the papers studied we found that the 4 evaluation metrics used to assess the proposed models are Recall, Accuracy, Precision and Specificity.

\hspace{-0.55cm}\textbf{Specificity and Sensitivity}\\
Specificity and sensitivity are standardized and recognized measures for evaluating performance in medical image segmentation. Sensitivity, also known as recall or true-positive rate, focuses on the capabilities of detecting true positives in pixel classification, while specificity, also known as the true-negative rate, evaluates the capabilities of correctly identifying true-negative classes (such as the background class). Specificity in image segmentation tasks demonstrates the ability of the model to recognize the background class in an image. Specificity ratios near 1 are expected due to the large percentage of pixels labeled as background compared to the region of interest (ROI). Therefore, specificity is a good statistic for checking the model's operation, but not so much for its performance.\\

\hspace{-0.55cm}\textbf{Accuracy / Rand Index} \cite{bib151}\\
Accuracy, also known as the Rand Index or Pixel Accuracy, is defined as the number of correct predictions, including positive and negative predictions, divided by the total number of predictions. However, in the case of unbalanced data, accuracy may not be the best measure of performance \cite{bib28, Ekert2019DeepLF}. Measuring accuracy will always produce an unrealistic result due to the inclusion of true negatives. Even when predicting the segmentation of an entire image as a background class, accuracy rates are often greater than or close to 100\%. Therefore, the misleading accuracy measure is not suitable for evaluating medical models based on the deep learning approach.\\

\hspace{-0.55cm}\textbf{F-measure, IoU, and DSC}\\
The F-measure, often known as the F-score, is one of the most extensively used metrics for quantifying performance in computer vision. It is derived from the sensitivity as well as the precision of a prediction, which analyzes the overlap between the expected segmentation and the ground truth. However, by considering accuracy, it also penalizes false positives, which are prevalent in datasets with significant class imbalance. There are two prominent metrics based on the F-measure: intersection over union (IoU), often known as the Jaccard index or Jaccard similarity coefficient, and the Dice similarity coefficient (DSC). DSC is referred to as the harmonic mean of sensitivity and precision. The IoU penalizes under- and over-segmentation more than the DSC. Even though these scores are suitable metrics and the most common metric in validating medical image segmentation \cite{bib23}, they are not employed for evaluating the performance of the suggested models in a majority of the articles analyzed in our SLR.\\

\hspace{-0.55cm}\textbf{ROC and AUC}\\
The ROC curve, short for Receiver Operating Characteristic, is a line plot that visualizes a classifier's performance with different discriminating thresholds. The performance is measured by comparing the true positive rate (TPR) to the false positive rate (FPR). In the medical field, ROC curves are widely accepted as a standard metric for comparing several classifiers when assessing diagnostic tests. Hanley and McNeil \cite{bib211} proposed the area under the ROC curve (AUC) as a single-value performance assessment for diagnostic radiology. The AUC measure is becoming a popular tool for validating machine learning classifiers. It should be noted that an AUC has a value ranging from 0 to 1. A model with 100\% incorrect predictions has an AUC of 0.0, while one with 100\% correct predictions has an AUC of 1.0.\\

\hspace{-0.55cm}\textbf{Other metrics}\\
For a segmentation system to be helpful and make a significant contribution to the area, its performance must be systematically tested. Furthermore, the evaluation must be carried out using conventional and well-known measures that allow for fair comparison. Other metrics exist and can be used based on the research topic and interpretive emphasis of the study. We refer to the excellent studies of Taha et al. \cite{bib212}. Additionally, Wang et al. \cite{bib213} describe the application of metrics in the evaluation of natural, medical, and remote sensing images and present the supervised and unsupervised evaluation methods in the application.

\subsection{Public Datasets for Dental Radiography Analysis}
Dental studies heavily rely on diverse datasets for optimal performance. Addressing the last research question (RQ7), our investigation, as summarized in Table \ref{tab9}, underscores that the majority of studies predominantly utilized private datasets. Remarkably, only tree studies opted for public datasets. Among the dataset types, panoramic radiographs emerged as the most frequently employed, providing a comprehensive single-film view of the maxilla and mandible, crucial for analyses such as orthodontic treatment and tooth growth. Additionally, a subset of studies utilized Cone-beam computed tomography (CBCT) for detailed three-dimensional (3D) imaging, while other types, including periapical radiographs and bitewing radiographs, offered vital diagnostic information for various dental conditions.

Our systematic literature review underscored a significant dearth of publicly available reference datasets suitable for extracting pertinent features related to oral medical conditions. Only three datasets were identified in the public domain \cite{bib215}, \cite{bib33}, and \cite{Brahmi2023}. Recognizing the pivotal role of data in influencing the performance of machine learning (ML) and deep learning (DL) models, the exploration of publicly available datasets, even in limited quantity, offers immense potential to enhance the robustness and generalizability of deep learning models in dentistry.

In the context of this scarcity, we would like to provide more detailed insights into specific publicly available reference datasets, each contributing uniquely to dental research:\\
\begin{itemize}
    \item The MLUA dataset, designed for dental caries segmentation in oral panoramic images, provides full-size X-ray images, corresponding segmentation masks, cropped images at tooth borders, and bounding box annotations. Accessible at\url {https://github.com/Zzz512/MLUA}, this resource supports research on accurate lesion detection, and its code and test samples are available for implementation.
    \item Panetta et al \cite{panetta}, introduces The Tufts Dental Database comprises 1000 panoramic dental radiography images with expert annotations, facilitating classification based on various criteria. Notably, it includes the radiologist's expertise through eye-tracking and think-aloud protocol recordings. This dataset serves as a benchmark for state-of-the-art systems, aiming to propel advancements in AI-driven abnormality detection, tooth segmentation algorithms, and the integration of radiologist expertise into AI applications.
    \item Dental Radiography Dataset, packed in a zip file, includes 1269 .jpg images along with three accompanying CSV files. Access the dataset at \url{https://www.kaggle.com/datasets/imtkaggleteam/dental-radiography/data}.
    \item the authors \cite{zhang2023n} collected dental panoramic radiographs from 106 pediatric patients (2 to 13 years old). Using EISeg and LabelMe, they created the world’s first dataset for caries segmentation and dental disease detection. Additionally, 93 pediatric radiographs were combined with three adult datasets (2,692 images) for a comprehensive segmentation dataset suitable for deep learning
    \item Lopez et al \cite{lopez} introduces a database comprises 598 panoramic radiographs with dimensions of 2041 x 1024, available in JPEG format. The annotations for these panoramic radiographs are accessible at \url{]https://www.kaggle.com/datasets/humansintheloop/teeth-segmentation-on-dental-x-ray-images}
\end{itemize}

By highlighting these specific datasets, despite their limited number, we emphasize their critical importance in developing more reliable and applicable models. Access to these diverse datasets from different centers continues to play an essential role in cross-center validation, ensuring the models' performance in various clinical contexts. This approach, encouraging cross-center collaboration and promoting generalization, remains essential to advance the field of deep learning in dentistry.
\subsection{How can future studies build upon the
findings of the reviewed papers to contribute novel insights or address existing limitations in the field?}
This section explores the gaps revealed through the analysis of fundamental studies in our systematic literature review, specifically focusing on the segmentation of dental panoramic images. By highlighting the current deficits, our objective is to establish a clear framework for future research in this specific domain. This critical synthesis of gaps will serve as a robust foundation for discussing potential directions in future research, paving the way for innovative contributions to enhance and enrich dental panoramic image segmentation. Following this, we will briefly present the main identified gaps and envisioned avenues for future research. The prevalent gaps identified in the literature review entail challenges linked to diverse facets. These challenges are delineated in the subsequent table \ref{tab8}.
\begin{longtable}{|p{2cm}|p{8cm}|p{4cm}|}
\caption{Challenges, Gaps and Future Work }
\label{tab8}
\\
\hline 
\textbf{Aspects}& \textbf{Limitations and studies} & \textbf{Improving Solutions and Charting Future Paths }\\
\hline
Dataset Size and source& 
Dataset size is pivotal in deep learning model development as it directly impacts the model's capacity to learn and generalize. A larger dataset provides more diverse examples, enabling the model to capture a broader range of patterns and variations present in real-world scenarios.
 \newline $-$ The dataset used was relatively small \cite{bib22, bib25, bib36, Lee2021DeepLF, You2020DeepLD,Tian2019AutomaticCA}. 
\newline$-$ To assess tooth-level or case-level Residual Bone Loss (RBL) for a large cohort, the model requires more images for further optimization.\cite{bib16}
\newline$-$ The dataset is collected from a single organization or single device , limiting its diversity \cite{bib18, bib19, bib23, bib25, bib31, bib35, Krois2021GeneralizabilityOD, Baaran2021DiagnosticCO, Lee2021DeepLF}.
\newline$-$ Absence of external test group \cite{Baaran2021DiagnosticCO, Cantu2020DetectingCL}.
\newline $-$ Heterogeneity in Image Sources \cite{amasya2023} or Distribution of Heterogeneous Images \cite{bib28}.
\newline $-$ The data collected for the study were unbalanced due to high labeling costs. \cite{bib20}
&$-$Employing data augmentation methods \cite{bib22, bib30}.
\newline$-$ Exploring Fine-tuning with transfer learning \cite{bib29, Setzer2020ArtificialIF}.
\newline$-$ Enriching the dataset with diverse data from multiple global sites and from various devices \cite{bib18, bib19}.
\newline$-$ Exploring the integration of non-imagery data, such as patient history, clinical signs, and symptoms, with imagery data within a deep learning system \cite{bib25, bib28, bib31}.
\newline$-$ Seek external validation of the neural network on diverse test sets to assess generalizability.
\newline Amasya et al. (2023)\cite{amasya2023} showcased DiagnoCat AI's adaptability across diverse image sources and configurations, independent of specific devices or imaging parameters \url{https://diagnocat.com}.\\
\hline
Image Quality&The image quality defined by attributes such as clarity, sharpness, and minimal noise. 
\newline $-$  Poor-quality images, such as those with overlapping teeth or distorted tooth length, can mislead the diagnosis, and the model does not automatically exclude them \cite{bib16, bib36}.
\newline $-$ Capturing various aspects of a tooth and the difficulties associated with internal tooth structure and the interproximal tooth surface in intraoral photography \cite{bib27, bib32, Chen2021HierarchicalCO, amasya2023}.
\newline $-$ Handling Incorrectly Oriented Radiographs \cite{bib32}.
\newline $-$ CBCT provides high resolution, but its accuracy in density measurement is inferior to traditional CT \cite{}
& $-$ Explore methods for preprocessing or filtering poor-quality images before analysis.
\newline $-$ Improve the model's robustness to handle variations in image quality.
\newline $-$ Conduct an investigation into a more realistic simulated X-ray acquisition process \cite{bib24}.
\\
\hline
Data Processing and Labeling & Refines raw data through operations like resizing and normalization, optimizing it for effective model training. Simultaneously, data labeling assigns categorical or numerical tags to each data point, providing crucial ground truth for supervised learning and enhancing the model's ability to generalize and make accurate predictions.
\newline $-$ Manual Mask Creation and Pre-processing poses a challenge due to its time-consuming nature and the need for expertise, consequently restricting scalability and potentially introducing errors \cite{bib19}.
\newline $-$ In the benchmark dataset, certain categories are labeled more coarsely than others \cite{Zhao2020TSASNetTS}.
\newline $-$Complexities in Labeling: especially when premolar teeth are missing without clear spacing. \cite{ali2023}
& $-$ Investigate automated or semi-automated methods for mask generation. Explore unsupervised learning techniques for region identification \cite{bib19}.
\newline $-$ The ambiguity in bounding box boundaries persists in object detection \cite{bib29}.\\
\hline
Not a Replacement for Clinical Data & While advantageous, deep learning models are not designed to replace the vital clinical data and expertise provided by healthcare professionals.
\newline $-$ The resulting model has not been used in a clinical setting, and it is not designed to replace periodontal charting and other clinical data.\cite{bib16, bib31, bib34, Cantu2020DetectingCL}.
\newline $-$ The study acknowledges a low success rate in identifying dental caries and calculus, making the AI models impractical for clinical use \cite{Baaran2021DiagnosticCO}.
\newline $-$ 
\newline $-$
&$-$ Explore seamless integration of the model's predictions with periodontal charting and data.\\
\hline
Specific limitation& Additional factors that impact the development, deployment, and effectiveness of deep learning models.
 \newline $-$ The model faces challenges in accurately identifying tooth numbers or positions, particularly when multiple teeth are missing \cite{bib16}.
\newline $-$ The model faces difficulties in precisely localizing specific dental issues.\cite{bib18, bib19}
\newline $-$ Increased Complexity and Training Time \cite{bib20}.
\newline $-$ The study did not compare the diagnostic performance of CNNs with that of dentists \cite{bib22}.
\newline $-$ Image Dimensions and platform Memory Constraints (Google Colab) \cite{bib28, brahmi2023automatic}.
\newline $-$ Overfitting \cite{bib30}.
\newline $-$ While capable of estimating age and sex with a single tooth, the models' performance is anticipated to enhance with multiple teeth, potentially limiting applicability in scenarios with few visible or accessible teeth \cite{bib34}.
\newline $-$ Exemplary Use-Case Focus \cite{Krois2021GeneralizabilityOD}.
\newline $-$ Determining the estimation success of each type of tooth (incisors, canines, premolars, molars).
\newline $-$ The Faster R-CNN network, while generally successful, encountered issues in recognizing two 'half teeth' as a single intact tooth due to the inherent limitations of Convolutional Neural Networks (CNNs) not considering spatial relationships between image features \cite{Chen2019ADL}.
& Distinguishing Implants with Different Internal Connection Types \cite{bib24} .
 \newline $-$ Understanding the trade-offs between image resolution and memory constraints \cite{bib28}.
 \newline $-$ Dropout \cite{bib30}. 
 \newline $-$ Investigating the impact of lower radiation doses on AI performance in dental imaging.
 \newline $-$ Automated image analysis based on 3D imaging to assist clinicians in lesion detection and contribute to the differential diagnosis of periapical pathology and nonodontogenic lesions \cite{Setzer2020ArtificialIF}.\\
\hline
\end{longtable}
\section{Limitations of the Study}\label{sec7}
In this systematic literature review (SLR), various deep learning-based techniques for dental image segmentation were identified. However, there are a few limitations of our SLR that should be taken into consideration. These limitations are listed as follows:
\begin{itemize}
\item Only journal articles discussing dental image segmentation were included in this SLR. Through our search technique, several unrelated research publications were discovered and excluded from this review during the preliminary stages of the study. This ensured that the selected research papers aligned with the investigation's requirements. However, it is believed that including additional sources such as books or conference papers could have enhanced the review.
\item This review is limited to papers available in the English language.
\item It is possible that other digital libraries with relevant studies were overlooked, although digital databases were considered when reviewing the study articles.
\item The effect of hyperparameters such as the number of epochs, batch size, and the number of layers on model performance has not been addressed in this study.
\end{itemize}
\section{Conclusion and future work}\label{sec8}
Over the past decades, there has been a notable increase in research interest at the intersection of deep learning and oral healthcare analysis, leading to continuous advancement in innovative models and methodologies year after year. These algorithms have the capability to assist dentists in various aspects of their clinical practice, including precise tooth identification, detection and categorization of dental pathologies, assessment of prosthetic structures, and error reduction. Deep learning algorithms hold considerable potential for streamlining the clinical workflows of dentists, thereby resulting in improved treatment outcomes.

The present study conducted a meticulous review of noteworthy articles focusing on deep learning methodologies applied to dental image segmentation. We systematically categorized these recent publications based on their specific objectives and conducted an in-depth exploration of the predominant evaluation metrics commonly used. Additionally, we addressed some of the significant challenges that influence the performance of these models. We anticipate that this study will provide readers with a comprehensive understanding of the essential dimensions of this field, highlight the most significant advancements, and shed light on potential avenues for future research.

In our forthcoming research endeavors, we aspire to capitalize on the insights gained from this rigorous systematic literature review (SLR). Our primary goal is to design a novel algorithm for dental pathology segmentation by harnessing the capabilities of deep learning. Importantly, we will also prioritize the interpretability of deep learning methodologies to mitigate their inherent opacity. Given the dynamic nature of this field, we envisage that this article will take the form of an ongoing review, with regular updates to reflect the emergence of new publications in this domain.\\
\section*{Declarations}
\textbf{- Conflicts of Interest} \\
The authors declare that they have no known competing financial interests or personal relationships that could have appeared to influence the work reported in this paper.\\
  \textbf{- Ethical Approval } \\
Not applicable \\
  \textbf {- Availability of supporting data}  \\
  Data will be made available on request.\\
 \textbf {- Competing interests} \\
There are no conflicts of interest to report \\
 \textbf {- Funding} \\
Not applicable \\
 \textbf {- Authors' contributions} 
\begin{itemize}
    \item  All authors conceptualized and wrote the main manuscript text.
    \item  Imen jdey conceptualized all tables and figures.
    \item Walid Brahmi prepared all figures and tables . 
    \item Adel Alimi reviewed the final version of the manuscript.
    \item All authors approved the final version of the manuscript.
\end{itemize}
 \textbf{- Acknowledgments }\\
This work was supported by the REGIM-Lab (Research Laboratory in Intelligent Machines, code LR11ES48) at the National Engineering School of Sfax (ENIS), University of Sfax, and the Faculty of Science and Techniques of Sidi Bouzid, University of Kairouan, Tunisia."
%\begin{appendices}

%\end{appendices}
\leavevmode

\hfill
\begin{landscape}
\begin{longtable}{|p{1.5cm}|p{4.5cm}|p{1cm}|p{5cm}|p{2cm}|p{5cm}|}
\caption{list of reviewed studies.}
\label{tab9}
\\
\hline 
\footnotesize \textbf{Authors}&\footnotesize \textbf{Dataset characteristics}&\footnotesize \textbf{Data Augmentation? }&\footnotesize \textbf{Main objective\& Architecture} &\footnotesize \textbf{Task} &\footnotesize \textbf{Performance} %(Best Value)%}
\\ \hline 
\footnotesize Brahmi et al\cite{brahmi2023automatic}, 2023&\footnotesize  \textbf{A public dataset} consisting of 107 panoramic x-ray images\cite{Brahmi2023}.&\footnotesize No&\footnotesize 
The Mask-RCNN network was utilized to perform pixel-level instance segmentation on a dataset1 containing 107 panoramic X-ray images of both young and adult patients. Subsequently, the Mask-RCNN network underwent training on dataset2, which comprises 73 panoramic X-ray images excluding those of young patients. & \footnotesize Instance segmentation and  Detection &\footnotesize mAP = 90\%, F1-sore = 63\% and precision = 96\% 
\\ \hline 
\footnotesize Choi et al\cite{choi2023}, 2023&\footnotesize  Private dataset consisting of 402 periapical radiography (PA), 138 Dens evaginatus and 264 normal cases. Divided into training set (N = 282, 70\%), validation set(N = 70, 18\%) and test set (N = 50, 12\%).&\footnotesize  
Yes&\footnotesize Five  DL models in image classification, including a simple convolutional neural network (CNN) model, VGG, DenseNet, ResNet, abd InceptionResNetV2, were selected for the diagnosis of Dens evaginatus (DE).&\footnotesize  Classification &\footnotesize F Model means (Full Image) and C Model mans (Cropped Image)
\newline AUC:	\newline
 F Model = 0.895,	C Model = 0.901
\newline Accuracy: \newline F Model  = 0.828, C Model = 0.832
\newline Precision: \newline F Model = 0.869, C Model = 0.856
\newline Recall:  \newline F Model = 0.871,	C Model = 0.898
\newline F1 Score: \newline F Model = 0.869,	C Model = 0.876
\newline Cohen’s Kappa: \newline F Model = 0.774,	C Model = 0.684
								
\\ \hline 
\footnotesize Chun et al\cite{Chun2023}, 2023&\footnotesize
A Private dataset included CBCT images from 50 patients, which were divided into a training set (64 volumes, 3546 images) and a test set (36 volumes, 1804 images). Each axial image, cropped to 512$*$512 pixels, served as input for the segmentation network. &\footnotesize No&\footnotesize Development and implementation of a distance-aware network for the automated classification of the three-dimensional (3D) positional relationship between an impacted mandibular third molar (M3) and the inferior alveolar canal (MC) in cone-beam CT (CBCT) images,  incorporating the accurate segmentation capabilities provided by the Dense121 U-Net. & \footnotesize Segmentation \ Classification &\footnotesize 
Segmentation performance:
\newline
IoU = 0.872, DSC = 0.920, Precision = 0.946, Recall = 0.918
\newline Classification performance: \newline
Accuracy = 1.00, Sensitivity = 1.00, Specificity = 1.00, AUC = 1.00
\\ \hline
\footnotesize  Amasya et al\cite{amasya2023}, 2023&\footnotesize  Private dataset consisting of 6000 panoramic radiographs with 45,161 annotated instances.&\footnotesize No&\footnotesize Two separate models are trained for different tasks. The first model uses Mask R-CNN with a pretrained ResNet-101 backbone to detect teeth, segment their masks, and define their numbering. The second model, based on the Cascade R-CNN architecture, is used for predicting periodontal bone loss. & \footnotesize Segmentation \newline Classification &\footnotesize For tooth conditions:
Overall F-score, accuracy, and Cohen's kappa coefficients were found to be 0.948, 0.977, and 0.933 for the binary results.
\newline For multiclass results, the corresponding values were 0.992, 0.988, and 0.961.
\newline For bone loss detection:
Overall F-score, accuracy, and Cohen's kappa coefficients were found to be 0.985, 0.980, and 0.956 for the binary results.
\newline For multiclass results, the corresponding values were 0.996, 0.993, and 0.974.\% 
\\ \hline
\footnotesize  Ali et al\cite{ali2023}, 2023&\footnotesize Private dataset, initially comprising 3818 panoramic radiographs, was refined to 3138 images, with Experiment 1 utilizing 2615 for training and 523 for testing, while Experiment 2 employed a six-fold cross-validation on the entire set, essential for evaluating tooth and prosthesis detection models. &\footnotesize Yes &\footnotesize A novel method using two YOLOv7-based object detectors to enhance teeth detection in dental panoramic X-rays, addressing challenges posed by prosthetic treatments. The approach involves detecting tooth and prosthesis candidates, assigning approximate tooth numbers to prosthetic candidates, and optimizing the results, significantly improving detection efficiency. & \footnotesize Detection \newline Classification &\footnotesize Mean Average Precision (mAP) for tooth detection = 0.982.
\newline Mean Average Precision (mAP) for prosthesis detection = 0.983.
\\ \hline
\footnotesize Lee, C. et al\cite{bib16}, 2022 &\footnotesize Private dataset. 693 periapical radiographic images , divided to 70\%, 10\%, and 20\% for training, validation, and testing.&\footnotesize No& \footnotesize The task was to develop and evaluate different variations of the U-Net architecture for the segmentation of three different areas: the bone area, tooth, and cemento-enamel junction (CEJ) line. &\footnotesize Segmentation & \footnotesize bone area  DSC = 0.96 , JI = 0.93 and PA = 0.96 \newline tooth shape DSC = 0.95 , JI = 0.91  and PA = 0.89  \newline CEJ line segmentations DSC = 0.91  , JI = 0.88  and PA = 0.99  %&Images with poor quality and When multiple teeth are missing, the model is not able to accurately identify the tooth number (position)
\\ \hline 
\footnotesize Imak, A  et al \cite{bib17}, 2022&\footnotesize   Private dataset. 380 periapical images. &\footnotesize No&\footnotesize  A score-based multi-input CNN ensemble (MI-DCNNE) for the automatic diagnosis of dental caries & \footnotesize  classification &\footnotesize accuracy = 99.13, sensitivity = 98\% , specificity = 100\%, precision = 100\% and  F1-score = 98.99\% 
\\ \hline  \footnotesize Zhang, X. et al \cite{bib18}, 2022 & \footnotesize Private dataset. 3932 oral photographs. Divided into training set (N = 2507, 63.76\%), validation set(N = 300, 7.63\%) and test set (N = 1125, 28,6\%). & \footnotesize Yes& \footnotesize Dental caries DL model adapted from Single Shot MultiBox Detector (SSD)   &\footnotesize  Segmentation \ Classification& \footnotesize ROC = 85.65\%. %& limited dataset 
\\ \hline  \footnotesize Oztekin, F. et al \cite{bib19},2022&\footnotesize Private dataset. 250 panoramic images of 2048×1024px. Randomly divided into 70\% for training, 20\% for validation, and 10\% for testing &\footnotesize No & \footnotesize A U-Net-based deep learning model for the automatic detection and classification of amalgam and composite fillings in panoramic images. &\footnotesize Segmentation  & \footnotesize mean IoU = 0.767 \newline Pixel Accuracy = 99.81\% %& Images were captured using a single device. %\newline %Pre-processing and manual mask production take a lot of time.
\\ \hline \footnotesize Park, J. et al \cite{bib20}, 2022& \footnotesize Private dataset.1414 panoramic X-ray&\footnotesize No&\footnotesize  a five-axis-based tooth recognition model using Faster R-CNN &\footnotesize Segmentation &\footnotesize  IoU = 72\% %& unbalanced dataset %\newline high labeling cost.
\\ \hline  \footnotesize Franco, A. et al \cite{bib21}, 2022&\footnotesize  . 4003 panoramic radiographs. divided into the age groups “under 15 years" (n = 2,254) and “equal or older 15 years" (n = 1,749). &\footnotesize Yes& \footnotesize Eight CNN architectures was used InceptionV3, Xception, Inception ResNetV2, ResNet50, ResNet101, MobileNetV2, VGG16 and DenseNet121 DenseNet121 to distinguish females and males using dentomaxillofacial features from a radiographic image &\footnotesize Segmentation \ Classification &\footnotesize DenseNet121- from scratch \newline Accuracy = 71.64\%,  F1-score = 70.64 \%, Precision = 71.59\%, Recall = 69.8\%, Specificity = 87.47\% 
\newline DenseNet121-transfer learning. \newline Accuracy = 82.18\%, F1-score = 8.064\% , Precision=0.8072\% , Recall= 0.8056\%,  Specifcity =0.9220\% 
\\ \hline \footnotesize Bayraktar, Y. et al \cite{bib22}, 2022 &\footnotesize Private dataset. 1000 bitewing radiographic divided into training (80\%) and testing-validation (20\%) groups &\footnotesize Yes& \footnotesize Modified (YOLO) model to detect caries lesions &\footnotesize Segmentation \ Classification &\footnotesize accuracy = 94.59\%, sensitivity = 72.26\%, specificity = 98.19\%and  AUC = 87.19
\\ \hline  \footnotesize Liu, M. et al \cite{bib23}, 2022 &\footnotesize  Private dataset. 254 CBCT. Divided into three subsets: the training set (154, 60.6\%), the validation set (30, 11.8\%), and the test set (70, 27.6\%) & \footnotesize Yes& \footnotesize U-Nets used to mandibular third molar and mandibular canal segmentation \newline ResNet-34 used tomandibular third molar and mandibular canal relation classification: 3 classes &\footnotesize Segmentation \ Classifcation& \footnotesize segmentation of mandibular third molar: mean Dice similarity coefcient (mDSC) = 97\% and a mean intersection over union (mIoU) = 96.06\%; \newline segmentation of mandibular canal: mDSC = 92\% and a mIoU of 90\%. \newline classification: sensitivity = 90.2\%, \newline specificity = 95.0\%, \newline accuracy = 93.3\% %&Images are collected from only one CBCT installation with fixed parameters.
\\ \hline \footnotesize Kohlakala, A. et al \cite{bib24}, 2022&\footnotesize  Private dataset. 483 simulated X-ray images, which contain implants inserted into either human or pig jaws of size 512×512 and saved in JPEG format. &\footnotesize Yes &\footnotesize Two fully convolutional networks model FCN1 and FCN2 used to dental implant detection and recognition & \footnotesize Segmentation \ Classification &\footnotesize segmentation accuracy = 94.0\% \newline classification accuracy = 71.7\%%& 
\\ \hline \footnotesize Moidu, N. et al \cite{bib25}, 2022&\footnotesize  Private dataset. 3540 periapical root areas (PRA). Divided into two subsets: the training and validation set (3000 PRA) and the test set (540 PRA) &\footnotesize Yes& \footnotesize A CNN model based on YOLO.v3 architecture to score the periapical lesion of mandibular teeth &\footnotesize Segmentation \ Classification &Segmentation: \footnotesize Dice coefcient = 89\% mean intersection over union (Jaccard index) = 90\%, recall = 89\% \newline precision = 90\% \newline Classification: sensitivity = 92.1\% \newline, specificity = 76\%, \newline accuracy = 86.3\%, \newline F1 score = 89\% \newline Matthews correlation coefficient (MCC) = 71\%. %&small sample size 
\\ \hline  \footnotesize Zhu, H. et al \cite{bib26}, 2022&\footnotesize Private dataset. 1159 panoramic images.Divided into three subsets: the training set (900), the validation set (135) and the test set (124). &\footnotesize No&\footnotesize CariesNet: an automated system for caries diagnosis that employs a U-shape encoder-decoder framework, a reverse attention mechanism, and a Res2Net backbone to achieve precise segmentation. &\footnotesize Segmentation&\footnotesize Dice coefficient = 93.64\%,accuracy = 93.61\%, precision = 94.09\% and recall = 86.01\%,
\\ \hline  \footnotesize Park, E. et al \cite{bib27}, 2022&\footnotesize Private dataset. 2348 intraoral photographic randomly assigned to training (1638), validation (410), and test (300) datasets&\footnotesize No&\footnotesize  Deep learning algorithm was utilized to perform tooth surface segmentation using U-Net, caries classification using ResNet-18, and lesion localization using Faster R-CNN. These techniques are applied in the context of caries detection..&\footnotesize segmentation \newline Classification &\footnotesize segmentation accuracy = 81.39\%, AUC = 83.7\%,\newline classification accuracy =9 2.5\%, sensitivity = 89.0\% and precision = 87.4\% %Internal tooth structure and the interproximal tooth surface cannot be captured in intraoral photography 
\\ \hline  \footnotesize Alotaibi, G. et al \cite{bib28}, 2022& \footnotesize Private dataset. 1724 periapical images. Divided in a training dataset (n = 1206; 70\%), a validation dataset (n = 345; 20\%), and a test dataset (n = 173; 10\%).& \footnotesize No & \footnotesize A CNN-based model VGG-16 to detect periodontal bone loss and classify the alveolar bone levels in teeth affected by periodontal disease. &\footnotesize classification& \footnotesize binary classification accuracy = 73.04\% , precision = 70\%, recall = 70\% and  Matthews correlation coefcient (MCC) = 51\% \newline multi-classification: accuracy = 59.42\%, precision = 83\%, recall =70\% and  Matthews correlation coefcient (MCC) = 0.65 
 %imbalanced image classes. 
\\ \hline  \footnotesize Jang, W. et al \cite{bib29}, 2022& \footnotesize Data available on request from the authors. 300 periapical radiographic image. split into 80:20 ratio (train / test). & \footnotesize No & \footnotesize Evaluate the diagnostic performance of the Faster R-CNN model with ResNet 101 backbone for the tasks of object detection, classification, and localization. &\footnotesize Detection \newline Classification \newline Localization & \footnotesize classification performance: \newline precision = 97.7\%, recall = 99.2\% and F1 score = 98.4\% \newline Segmentation performance  \newline IoU=90.7\%. 
\\ \hline \footnotesize  Mohammad, N. et al \cite{bib30}, 2022& \footnotesize Private dataset. 240 panoramic images.& \footnotesize Yes& \footnotesize Dynamic programming of active contour (DP-AC) and convolutions neural network model to segment the maturity development of the mandibular premolars. &\footnotesize segmentation \newline Classification& \footnotesize Classification : Cohen’s Kappa = 58\%, \newline accuracy = 77\% %&limited dataset
\\ \hline  \footnotesize Jiang, L. et al \cite{bib31}, 2022&\footnotesize Data available on request from the authors. 640 panoramic images. separated into a training set (80\%) and a test set (20\%) &\footnotesize Yes& \footnotesize UNet and YOLO-v4 were used to train a deep learning model for comprehensively diagnosing and staging periodontal alveolar bone loss. &\footnotesize Segementation \newline Classification &\footnotesize accuracy = 77\%, Precision = 77\%, Sensitivity= 77\%, Specifcity = 88\% and F1-score = 77\%.  %&the research was not conducted in a real clinical environment.\newline limited dataset.
\\ \hline \footnotesize  Kabir, T. et al \cite{bib32}, 2022& \footnotesize A public repository contained 116 panoramic images \cite{bib33}. \newline 1240 intraoral radiographs obtained from a private database. &\footnotesize No&\footnotesize U-Net segmentation model with ResNet-34 as the backbone was utilized to accurately segment different structures and regions of interest in dental images, such as teeth, CEJ lines (Cementoenamel Junction), and the bone area. & \footnotesize segmentation \newline Numbering &\footnotesize Detection Precision=0.99 \newline Detection Recall = 99\% \newline Numbering Precision = 96\% \newline Numbering Recall = 96\% %& The proposed model cannot handle incorrectly oriented intraoral radiographs. \newline limited dataset. 
\\ \hline \footnotesize Vila-Blanco, N. et al \cite{bib34}, 2022& \footnotesize Private dataset. 1746 dental panoramic radiographs (orthopantomograms or OPGs) was split into training, validation, and test subsets,which contained 60\%, 20\%, and 20\% of the cases, respectively.&\footnotesize Yes&\footnotesize  XAS: Automatic yet eXplainable Age and Sex determination.&\footnotesize segmentation \newline Classification&\footnotesize Tooth segmentation \newline mAP@0.5 = 96.4\%, mAP@0.75 =94.4\% \newline Sex classification Accuracy = 91.8\%
\\ \hline  \footnotesize Bonfanti-Gris, M. et al \cite{bib35}, 2022 &\footnotesize Private dataset. 300 panoramic radiographs exported in jpeg file &\footnotesize No& \footnotesize  A pre-trained Convolutional Neural Network (CNN) called Denti.Ai to perform the tasks of detecting and classifying dental structures in panoramic radiographs. .&\footnotesize segmentation \newline Classifcation&\footnotesize Sensitivity = 56.92\% \newline Specificity = 92.20\%%&limited dataset 
\\ \hline \footnotesize Aljabri, M. et al \cite{bib36}, 2022&\footnotesize Private dataset. 416 panoramic radiographs &\footnotesize Yes & \footnotesize 4 deep learning models were developed to classify the type of canine impaction (Type I or Type II) from panoramic dental radiographic images: DenseNet-121, VGG-16, Inception V3, and ResNet-50. &\footnotesize Classification &\footnotesize Experiment 1 Accuracy (unbalanced data : 282 samples represent Type I and 134 Type II )\newline Inception V3 0.8095 \newline ResNet-50 0.7619 \newline DenseNet-121 0.7976 \newline VGG-16 0.6548 \newline Experiment 2 Accuracy (balanced data : 134 samples represent Type I and 134 Type II ) \newline Inception V3 0.9259 \newline ResNet-50 87.04\% \newline DenseNet-121 68.52\% \newline VGG-16 57.41\%. %& small size of the labeled dataset.
\\ \hline \footnotesize  Sivasundaram, S. et al \cite{Sivasundaram2021PerformanceAO}, 2021&\footnotesize Data available on request from the authors. 1171 panoramic dental images, augmented into 3513. &\footnotesize Yes& \footnotesize Modified LeNet  architecture for classifying the oral cyst images and a morphology-based segmentation method for segmenting the cyst regions in the classified cyst images&\footnotesize  segmentation \newline Classification& \footnotesize sensitivity = 98.3\%, specificity = 98.8\%, an accuracy = 98.5\%, precision =97.53\%, F1 score = 98.06\%, average DSC = 98.04 and IoU = 97.6. 
\\ \hline \footnotesize Hsiao, T. et al \cite{Hsiao2021DiseaseAM}, 2021& \footnotesize Private dataset. Optical coherence tomography images. 32000 B-scans for training, 8000 B-scans for validation data, and 8000 B-scans for testing &\footnotesize No &\footnotesize A VGG16-based automated model is employed to detect the presence of subgingival calculus and accurately identify the location of the lesion in optical coherence tomography (OCT) images.& \footnotesize Detection \newline Localization& \footnotesize accuracy = 95.06\%. \newline AUC = 0.973.
\\ \hline  \footnotesize Krois, J. et al \cite{Krois2021GeneralizabilityOD}, 2021&\footnotesize Two datasets each have 650 images available if needed within data protection regulation boundaries.&\footnotesize Yes & \footnotesize U-Net was used to evaluate the generalizability of deep learning models for the detection of apical lesions on panoramic radiographs and to identify methods for improving their performance. &\footnotesize segmentation& \footnotesize Model trained only on First dataset image :\newline F1-score = 54.1\% `\newline Model trained only on second dataset image :\newline F1-score =32.7\% \newline Cross-Dataset training \newline F1-score (First dataset) = 50.9\% \newline F1-score (Second dataset) = 46.1\%.
\\ \hline \footnotesize Basaran, M. et al \cite{Baaran2021DiagnosticCO}, 2021& \footnotesize Private dataset. optical coherence tomography images. 1084 dental panoramic radiographs. & \footnotesize No & \footnotesize AI model based on Faster R-CNN and Google Net Inception v2 to detect dental conditions. &\footnotesize segmentation& \footnotesize best sensitivity = 96.74\%, best precision= 92.59\% and the F1-score = 94.33\%.%&The success rate of identifying dental caries and calculus were too low\newline Only one device and standard parameters were used to image acquisitions.\newline No external data was used to evaluate the model’s performance.
\\ \hline  \footnotesize Chen, H. et al \cite{Chen2021DentalDD}, 2021&\footnotesize Private dataset. optical coherence tomography images. 2900 digital dental periapical radiographs &\footnotesize No &\footnotesize  Faster R-CNNs to detect diseases including decay, periapical periodontitis, and periodontitis in dental periapical radiographs.&\footnotesize Detection \newline  localization&\footnotesize decay\newline IoU = 0.71, Precision = 0.61, Recall = 0.54 and AP = 0.45 \newline periapi\newline IoU =0.69, Precision = 0.51, Recall = 0.51 and AP = 0.36 \newline periodo\newline IoU = 0.68, Precision = 0.56, Recall = 0.61 and AP =0.43.
\\ \hline \footnotesize  Lee, S. et al \cite{Lee2021DeepLF}, 2021 &\footnotesize  Private dataset. 304 bitewing radiographs. Divided into 149 radiographs for training on both structure and caries segmentation, 105 radiographs for training on caries segmentation, and 50 radiographs with no dental caries.&\footnotesize Yes &\footnotesize The U-Net architecture is used for the early detection of initial dental caries (tooth decay).&\footnotesize Detection & \footnotesize  precision = 63.29\%;\newline recall = 65.02\%; \newline F1-score = 64.14\%.%& small dataset
\\ \hline  \footnotesize Bilgir, E. et al, 2021&\footnotesize Data available on request from the authors. 2482 panoramic radiographs. Divided into training (80\%), validation (10\%), and test (10\%) groups. &\footnotesize No&\footnotesize Faster R-CNN Inception v2 model, was used to automatically detect and number teeth on the panoramic radiographs. &\footnotesize segmentation &\footnotesize Sensitivity (Recall) = 0.9559 \newline Precision = 0.9652 \newline F1Score = 0.9606.
\\ \hline  \footnotesize Chen, Q. et al \cite{Chen2021HierarchicalCO}, 2021&\footnotesize Private dataset. 175 full-jaw 3D tooth models&\footnotesize Yes&\footnotesize A basic CNN model for 8-class tooth type classification (first premolar, second premolar, first molar, second molar in maxilla and mandible respectively)&\footnotesize classification&\footnotesize accuracy = 91.35\%, \newline precision = 91.49\%, \newline recall = 91.29\%, \newline F1-score = 0.91
\\ \hline  \footnotesize Rajee, M. et al \cite{Rajee2021GenderCO}, 2021&\footnotesize  Private dataset. 1000 panoramic dental images divided into 600 training samples and 400 testing samples&\footnotesize No & \footnotesize Gender classification on digital dental x-ray images using deep convolutional neural network model based on Resnet50 architecture. &\footnotesize Segmentation \newline Classification& \footnotesize dice = 0.94 \newline accuracy = 98.27\newline Sensitivity = 98.04 \newline Specificity = 98.51 \newline Precision = 98.52 
\\ \hline  \footnotesize Singh, P. et al \cite{Singh2020NumberingAC}, 2020 &\footnotesize Private dataset. 400 panoramic dental images, divided into 240 training samples and 160 testing samples &\footnotesize Yes& \footnotesize Numbering and Classification of Panoramic Dental Images Using 6-Layer Convolutional Neural Network (DCNN) &\footnotesize Pre-processing \newline  Segmentation \newline Numbering  \newline classification (4 classes molar, premolar, canine and incisor).& \footnotesize Augmented database \newline accuracy = 95\% \newline Sensitivity = 98\% \newline Specificity = 92\% \newline \newline Original dataset \newline accuracy = 92\% \newline Sensitivity = 90\% \newline Specificity = 85\% 
\\ \hline \footnotesize You, W. et al \cite{You2020DeepLD}, 2020 &\footnotesize Data available on request from the authors. 984 intraoral divided into 886 training samples and 98 testing samples. &\footnotesize No&\footnotesize Deep learning-based dental plaque detection on primary teeth based on DeepLab and DeepLabV3 architectures . &\footnotesize segmentation &\footnotesize mean intersection-over-union (MIoU) = 0.726 %&\newline Small number of training photos
\\ \hline  \footnotesize Kwak, G. et al \cite{Kwak2020AutomaticMC}, 2020 &\footnotesize  Data available on request from the authors. 49094 CBCT images, divided into train:valid:test sets with the ratio of 6:2:2.&\footnotesize No & \footnotesize An Automatic mandibular canal detection based on 2D SegNet, 2D and 3D U-Nets architectures.& \footnotesize Segmentation &\footnotesize 2D U-Net accuracy = 82\% \newline 2D SegNet accuracy = 96\% \newline 3D U-Net accuracy = 99\%. 
\\ \hline \footnotesize Zhao, Y. et al \cite{Zhao2020TSASNetTS}, 2020 &A\footnotesize  benchmark public dataset. 1500 panoramic X-ray images divided into 1200 training samples, 150 validation samples and 150 testing samples.& \footnotesize No& \footnotesize A Two-Stage Attention Segmentation Network (TSASNet) on dental panoramic X-ray images &\footnotesize segmentation& \footnotesize Accuracy = 96.94 \newline Specificity = 97.81\newline Precision = 94.97\newline Recall = 93.77 \newline Dice = 92.72 %teeth segmentation has h some difficult cases: %The first one is the segmentation of the supernumerary tooth situation. because there is no distinct boundary between those teeth, and some teeth even overlap directly Another one is the segmentation of teeth with fillers. In this case, the tooth roots are almost homogenized by the upper jaw
\\ \hline  \footnotesize Cantu, A. et al \cite{Cantu2020DetectingCL}, 2020 &\footnotesize Private dataset. 3686 bitewing radiographs. The data was divided into a training (3293 images ), validation (252 images ) and test dataset (141 images)&\footnotesize No & \footnotesize A deep learning model based on U-net to detect caries lesions on bitewing radiographs&\footnotesize Segmentation&\footnotesize Accuracy 0.80 \newline Sensitivity 0.75\newline Specificity 0.83 \newline F1score 0.73 \newline MCC 0.57 

%& dataset cannot claim full generalizability\newline reference test was built on “fuzzy” labeling (the manifold of provided annotations from different annotators.)\newline\newline\newline
\\ \hline \footnotesize Setzer, F. et al \cite{Setzer2020ArtificialIF}, 2020 &\footnotesize  Private dataset. 20 CBCT images were divided into 16 CBCT images) were used for training, and 4 CBCT images was used for validation&\footnotesize Yes &\footnotesize A Deep Learning model for the automated segmentation of cone-beam computed tomographic (CBCT) images and the detection of periapical lesions. Based on a U-Net architecture.&\footnotesize Segmentation & \footnotesize DICE index \newline Lesion 0.52\newline Tooth structure 0.74 \newline Bone 0.78 \newline Restorative materials 0.58\newline Background 0.95 %& Limited data samples
\\ \hline \footnotesize Chen, H. et al \cite{Chen2019ADL}, 2019&\footnotesize Data available on request from the authors. 1250 periapical film divided into training (800), validation (20.), and test (250) groups &\footnotesize No&\footnotesize  A deep learning approach for automatic teeth detection and numbering based on faster R-CNN (A total of 32 teeth classes were required to be recognized in the X-ray images) &\footnotesize segmentation &\footnotesize  Numbering \newline precision = 0.98\newline recall = 0.98\newline Mean IOU = 0.91\newline \newline Numbering \newline precision = 0.91\newline recall = 0.91 %& the propsed model recognized two ‘half tooth’ as an intact tooth
\\ \hline\footnotesize  Al Kheraif, A. et al \cite{Kheraif2019DetectionOD}, 2019& \footnotesize Private dataset. 1500 panoramic images divided into training of 800 with testing of 7000& \footnotesize No& \footnotesize Histogram enhancement with a convolutional neural network (CNN) to detect dental diseases&segmentation \newline classification &\footnotesize  Segmentation : Accuracy = 0.912; Specificity = 0.967; Precision = 0.89; Recall = 0.92 and F1 score = 0.94 \newline classification Accuracy = 97.07\% 
\\ \hline  \footnotesize Ekert, T. et al \cite{Ekert2019DeepLF}, 2019 &\footnotesize Private dataset. A synthesized data set of 2001 tooth segments from panoramic radiographs &\footnotesize Yes& \footnotesize A deep convolutional neural networks (CNNs) to detect apical lesions (ALs). &\footnotesize segmentation& \footnotesize AUC = 0.85, sensitivity = 0.65 and specificity = 0.87 %& limited sensitivity is related to the low prevalence of apical lesions (ALs) in the data set,
\\ \hline \footnotesize Tian, S. et al \cite{Tian2019AutomaticCA}, 2019 & \footnotesize Private dataset. 600 dental models &\footnotesize No&\footnotesize An automatic segmentation and classification method for 3D dental model via 3D CNN. &\footnotesize Segmentation\newline classification : seven tooth types which are the central incisors, lateral incisors, canines, first premolars, second premolars, first molars and second molars.& \footnotesize classification accuracy in Level1 network is 95.96\%, the average classification accuracy in Level2 network is 88.06\%, and the accuracy of tooth segmentation is 89.81\% %&limited number of dental samples
\\ \hline \footnotesize Lee, J. et al \cite{Lee2018DetectionAD}, 2018 & \footnotesize Private dataset. 3000 periapical radiographic, divided into a training and validation dataset (n = 2400 [80\%]) and a test dataset (n = 600 [20\%]) &\footnotesize Yes& \footnotesize GoogLeNet Inception v3 for detection and diagnosis of dental caries on periapical radiographs. &\footnotesize Segmentation\newline classification& \footnotesize premolar accuracy = 89.0\% , molar accuracy = 88.0\% , both premolar and molar accuracy = 82.0\% \newline AUC = 0.917 on premolar, AUC = 0.890 on molar, and AUC = 0.845 on both premolar and molar models.\\
 \hline \hline
\end{longtable}
\end{landscape}
\vfill
\hfill
\bibliography{main}% common bib file
\end{document}